# Topic Modeling the *Hàn diǎn* Ancient Classics (汉典古籍)

Colin ALLEN[1,2], Hongliang LUO[3], Jaimie MURDOCK[2], Jianghuai PU[1], Xiaohong WANG[1], Yanjie ZHAI[1], Kun ZHAO[3]

[1] Department of Philosophy, School of Humanities and Social Sciences, Xi'an Jiaotong University, Shaanxi, China
[2] Cognitive Science Program, Indiana University, Bloomington, Indiana, USA
[3] Institute of Computer Software and Theory, School of Electronic and Information Engineering, Xi'an Jiaotong University, Shaanxi, China

Authors listed alphabetically

**Abstract:** Ancient Chinese texts present an area of enormous challenge and opportunity for humanities scholars interested in exploiting computational methods to assist in the development of new insights and interpretations of culturally significant materials. In this paper we describe a collaborative effort between Indiana University and Xi'an Jiaotong University to support exploration and interpretation of a digital corpus of over 18,000 ancient Chinese documents, which we refer to as the "Handian" ancient classics corpus (汉典古籍 or *Hàn diǎn gǔ jí*, i.e, the "Han canon" or "Chinese classics"). It contains classics of ancient Chinese philosophy, documents of historical and biographical significance, and literary works. We begin by describing the Digital Humanities context of this joint project, and the advances in humanities computing that made this project feasible. We describe the corpus and introduce our application of probabilistic topic modeling to this corpus, with attention to the particular challenges posed by modeling ancient Chinese documents. We give a specific example of how the software we have developed can be used to aid discovery and interpretation of themes in the corpus. We outline more advanced forms of computer-aided interpretation that are also made possible by the programming interface provided by our system, and the general implications of these methods for understanding the nature of meaning in these texts.

## Introduction and Context

The use of computers to support scholarship in the humanities reaches back over 50 years.[1] The first decades of the twenty-first Century have seen the acceleration of humanities computing, particularly in North America and Europe, with the field coalescing around the label "Digital Humanities" (DH). The recent growth of DH is the product of a feedback loop caused by several factors including: (1) the increasing availability of digitized materials, especially on the World Wide Web; (2) increased computer storage capacity and processing capacity; (3) advances in text modeling and visualization algorithms; (4) deepening understanding by scholars of the interpretive possibilities provided by computational methods; (5) funding commitments from government and private foundations; and (6) last, but not least, the growing perception among many young scholars and doctoral students that DH is an exciting area of inquiry and an important enhancement to their career prospects.[2] DH projects may concern themselves with many different media: text, images, audio, video, etc. However, our focus here is the analysis of written texts. Textual analysis constitutes the largest component of DH. This is largely because written language has been central to the construction and transmission of intellectual culture, and because of the relative ease with which text can be encoded and shared. These factors have resulted in enormous amounts of textual material recently becoming available.

As an example from category (1), increased availability of digitized materials, we highlight the HathiTrust (HT) digital library.[3] It started as a collaboration among major university research libraries in the United States, to digitally scan the books in their collections.[4] The page images from these books have been converted to text using optical character recognition (OCR) software. The HT collection now comprises over 14 million scanned volumes, equivalent to around five billion (5,000,000,000) pages based on the HathiTrust estimated average of 350 pages per book.[5] (Perhaps as many as half a million of these books are Chinese language volumes, as determined by a search at babel.hathitrust.org.) By any standard, this is a vast amount of text: more than could be read in multiple human lifetimes. Because of its enormous scale, the digitized pages in the HT are relatively uncurated. Despite the care with which editors prepared the original physically printed editions, the images and OCR representations of the pages contain scanning errors that have not been corrected. Nevertheless, the HT digital library is a treasure trove for DH that offers multiple possibilities for analysis.[6] At the same time, traditional scholarly editions have become increasingly digital, making available highly curated editions of historically and culturally significant text corpora. These projects use a labor-intensive process of inserting markup according to widely adopted standards for the semantic web, such as TEI (Text Encoding Initiative) and OWL (Web Ontology Language).[7] Only some of these editions are fully accessible to all users, however, because the costs of producing and curating them must often be recovered by subscriptions paid by libraries or individual users. For some scholarly purposes, the standard of care with which such editions are produced is essential, but access to imperfectly digitized texts, such as provided by the HathiTrust and by Project Gutenberg,[8] is adequate for many projects. However, even full access to relatively open archives such as the HathiTrust faces some restrictions because of complicated copyright issues which vary from country to country.

None of the large-scale repositories would have been possible without the exponential growth in computer capacity that has occurred since the advent of computing. Known as “Moore’s Law”, the doubling of computer speed and memory every 18 months for the same cost has produced supercomputers capable of processing terabytes of data, and personal laptops capable of storing and processing far more material than any single person could hope to read in several lifetimes. At the same time, the Internet and its hypertext offshoot, the World Wide Web, have made distributed repositories and cluster computing possible. This growth in speed, storage, and networking, has been accompanied by increasingly sophisticated algorithms for processing text and visualizing the results. The earliest efforts in humanities computing focused mainly on counting and localizing key words in texts. Other DH approaches applied network analysis to names, dates, places and other metadata such as citations, extracted from text. More recently, techniques for modeling full text contents have been introduced by computer scientists. Originally developed for the purposes of information retrieval, techniques such as latent semantic analysis (LSA), probabilistic topic modeling using latent Dirichlet allocation (LDA), and neural-network models of word embeddings, have been adopted within DH.[9]

Although differing in their details, what these methods have in common is their representation of documents and words as vectors within a multidimensional space. In some representations, the dimensions of the space correspond to words. In other representations, the dimensions may correspond to concepts, topics, or other abstractions from the data. Algorithms based on these vector representations are capable of identifying hidden (latent) factors in text. Such representations allow for interesting and meaningful measures of similarity among terms and documents, for example the cosine of the angle between vectors, or well-defined information-theoretic measures on the probability spaces. With such methods, DH scholars are beginning to move beyond counting words, to detecting and analyzing patterns of historical significance at cultural scale[10] and at the scale of an individual.[11]

There is a small but growing literature on large-scale statistical modeling of Chinese language texts. Ouyang analyzed a corpus of over 40,000 ancient documents downloaded from multiple sources. This was used to plot the temporal distributions of word frequencies and geographic distributions of authors.[12] Huang and Yu modeled the SongCi poetry corpus, first converting it to tonally marked pinyin to conserve poetically important pronunciation information.[13] Nichols and colleagues reported initial modeling of the Chinese Text Project corpus[14] in a conference paper. (Further below, we describe differences between this corpus and the Handian.) With additional collaborators, this group has now conducted two studies that are currently unpublished but under review. In the first, they apply topic models to address scholarly questions about the relationships among important texts of Ancient Chinese philosophy. In the second, they use topic modeling to investigate the concepts of mind and body in ancient Chinese philosophy.[15] Although we share similar scholarly objectives with these researchers, our approach in this paper is unique in that for the first time anywhere we bring the benefits of computational modeling of ancient Chinese texts to a robust public platform that is mirrored on both sides of the Pacific. Besides being just a useful portal to the texts, our approach foregrounds the interpretive issues surrounding topic models,[16] and makes more sophisticated exploration and analysis of interpretive questions possible for experts and novices alike.

The Chinese language presents interesting challenges for humanities computing. Both modern and ancient Chinese, but especially the latter, rely heavily on context for the interpretation of individual characters and words[17] and some researchers have argued that differences in Chinese morphology make some of the techniques that work well for DH work in Western languages less applicable to Chinese.[18] Words in Chinese are highly polysemous, requiring considerable amounts of context for their proper interpretation. The study of ancient Chinese philosophy is especially challenging because this ambiguity and openness to multiple interpretation seems to be deliberately exploited by the ancient masters.[19] Take, for example the character '道' which could refer to Taoism, but has up to 10 meanings in ancient Chinese texts, such as 'way' or 'road', and is also used as a verb to mean 'say'. At the same time, the long and relatively continuous history of the Chinese nation has enabled the transmission of a rich corpus of ancient texts to the present day. Computational modeling of these texts does not, as we see it, aim to remove the *human* from the *human*ities. Rather, by enabling the discovery and quantitative analysis of connections, computational methods promise at least these two benefits: (i) enhanced means of access to large sets of documents, and (ii) new sources of evidence about texts that can support the ongoing discussion of their interpretation relative to the past and the present. We are also interested in a more general theme (iii), concerning the potential broader significance for theoretical discussions of the nature of meaning and the role of language in conceptual schemes.

Our primary contribution in this paper is of type (i), to provide enhanced access to a corpus of ancient Chinese documents. Specifically, we introduce an application of the InPhO Topic Explorer[20] developed at Indiana University, Bloomington, USA, to a large, public corpus of ancient Chinese texts, resulting from collaboration with philosophers and computer scientists at Xi'an Jiaotong University, Shaanxi, China. We also discuss potential projects and future research of type (ii) concerning the analysis of the themes in ancient Chinese philosophy and other literary sources. We present a very brief discussion of the broader significance (iii) before the conclusions section of this paper.

## Selecting and Preparing the Corpus

A good understanding of Chinese intellectual culture during the classical period is important in itself, and essential for understanding the reception of Western ideas during various stages of China's history, and vice versa. As philosophers, we are particularly interested in philosophical texts, but we recognize that the boundaries between philosophy and other areas such as religion and political theory are fuzzy at best, and practically non-existent in some cultures or during certain periods of history. Thus, rather than try to demarcate "philosophy" from the rest, we decided to pursue our computational inquiry with as broad a corpus as we could locate.

A secondary consideration is that we want our work to provide a public benefit by being accessible to scholars and the public. It is less than optimal to analyze sources that only a few people -- not even all scholars -- have access to. For example, although the Wenyuange Edition of the Siku Quanshu archive[21] is of high quality for scholars, it is accessible only to those with subscriptions that are locked to specific IP addresses. Thus we conducted a scan of repositories of ancient Chinese documents, and found that the crowd-sourced website at zdic.net provided the best combination of quantity and access to a large number of classic texts, thanks to its permissive re-use policy under a Creative Commons 1.0 Public Domain Dedication.[22] The full website at www.zdic.net contains a dictionary of Chinese characters, a dictionary of words, dictionary of idioms and several other resources. Among them is the collection of classics identified as 汉典古籍 (*Hàn diǎn gǔ jí*) or Chinese classics— the portion we refer to as the "Handian" corpus — directly accessible at http://gj.zdic.net/, and it is this portion of the website that we chose to model. This section of the website is not without problems, however. It contains a diverse collection of different file formats, containing both traditional and simplified characters, and of varying quality because they have been crowd-sourced from many different users using many different sources, with varying degrees of scholarly care. A better-curated corpus is the Chinese Text Project used by Nichols, Slingerland and colleagues.[23] Although this site can be downloaded for private and academic use, its re-use policy is not as permissive as the Handian, and the online analysis tools require a subscription. Furthermore, because ctext.org is registered in Panama and hosted in the USA as well as directed towards English-speaking users, access by users in mainland China is generally slower and more difficult than zdic.net, which is registered and hosted in China.

For our initial goals, the benefits of accessibility, especially to Chinese users, outweighed the concerns about corpus curation quality. Such concerns are also partly mitigated by the topic modeling methods (described in more detail below). Because topic models treat documents as unordered "bags of words", they are relatively robust in the face of the "noise" provided by the variable quality of the texts. The techniques we describe here can be applied to more scholarly editions of the same texts. By demonstrating the power of the approach with the Handian corpus, we hope to encourage curators of scholarly editions to incorporate similar methods and make their efforts publicly available. We have made the products of our research available for all at our Indiana University website in the USA, mirrored at the Xi'an Jiaotong University website in China.[24]

In November of 2016 we crawled and downloaded the four sections of the Han classics from the gj.zdic.net site. These sections, which are derived from the Siku Quanshu (the library of the Qianlong Emperor in Four Sections) are the 经部 (*Jīng Bù*), containing Confucian classics, 史部 (*Shǐ Bù*), historiographic works, 子部 (*Zǐ Bù*), containing writings of the philosophical schools, and 集部 (*Jí Bù*), a section of miscellaneous anthologies, including poetry, drama, and other works of literature. Each of the sections contains a multi-level tree of further subsections terminating in text files. For example, within the 经 (*Jing*) section are three subsections, labeled 十三经 (thirteen classics), 十三经注疏 (thirteen classics annotations), and 经学史及小学类 ("history of classical studies and traditional Chinese philology"), and these are further subdivided.[25] We found that some of the files were index files listing the contents of the directories, so we discarded these.

We developed some custom mixture of automated and semi-automated methods to extract the original texts from the downloaded HTML pages. Next we cleaned the corpus by regularizing the characters and their encoding method. Because of the mixture of traditional and simplified characters in the corpus, we decided to map all characters to simplified characters. This entails a loss of visual, aural and etymological information, important for interpretation by knowledgeable readers, but of no direct use to the algorithms beneath the topic modeling process. (In the future we will provide additional support for both traditional and simplified characters within the Topic Explorer.)

After this preliminary processing, we found that quite a few files were empty — some representing documents lost to history, others not present for other reasons. So, we removed these files leaving 18,818 files for analysis.[26] These files contain approximately 100 million individual characters. Chinese does not use spaces to separate words, but some words comprise multiple characters. Hence, text modelers face a choice of whether to model the corpus character-by-character or whether to segment the text into words. Because the vast majority of ancient Chinese words are written as single characters, the character-by-character option may have been a reasonable choice for this corpus. It was our judgment, however, that segmentation of the texts into words rather than characters would improve interpretability of the models.[27] Software to address the word segmentation problem in modern Chinese exists, but these solutions are dictionary-based. Thus it was necessary for us to find and deploy a dictionary of ancient Chinese that we constructed from different sources.[28]

After applying the dictionary to our corpus, we identified nearly 84 million word tokens comprising nearly 85,000 unique word types. The most common word in the Handian corpus is 之 (*zhī,* it/this/for) at just over 1.25 million occurrences and the most common two-character word was 天下 (*tiānxià*, the World) at 83,805 occurrences, 93rd most frequent in the overall list. Very high frequency words are relatively uninformative and they tend to overwhelm the available methods for corpus analysis, both because of the additional time to process so many characters in a corpus of this scale, and because the highly frequent terms tend to dominate more meaningful terms in the trained models. Therefore, it is normal to develop a “stop list” of such words to remove them from the corpus.[29] Our stop list of 187 words is larger than the 132 words listed by Slingerland et al.,[30] and the two lists overlap in 50 words. The relative disjointness of the two can be explained by the differences in size and scope of the two corpora and the different objectives of the two projects. For example, we found it useful to filter out more of the frequently occurring number words.

## LDA Topic Modeling

Based on our previous experience working with large text collections within the InPhO team at IU, we chose to apply LDA (Latent Dirichlet Allocation) Topic Modeling to the Handian corpus. (LDA is named for the 19th C. mathematician Gustav Dirichlet who laid the foundation in probability theory for the technique.) LDA Topic modeling has become popular within DH in recent years, although the interpretation of this kind of model remains a matter of considerable discussion.[31] It treats documents initially as “bags of words” — that is, all grammatical structure and information about word order within sentences or documents is ignored, and the document’s initial profile is simply the frequency with which of all the words appear in it. Topic modeling aims to find latent (hidden) structure among these “bags of words”, by re-representing each document as a mixture of topics. A topic may also be thought of as a writing context, as we now explain.

We understand topic models to provide a theory about writing. Authors of documents combine different subjects of discussion. Different authors working within similar cultural contexts have overlapping interests in various subjects, but they combine the available topics differently. When writing about good behavior, for example, one may be concerned with the good behavior in the public sphere of business or politics or religion, or in the family or social community, or as a topic within moral philosophy. An author is more or less likely to use a given word when writing about each of these subjects. For example, the words ‘sister’ or ‘father’ are more likely to be used when the author’s subject is family than when writing about business. Other words may have very similar likelihoods of being used in these contexts. For example, the word “virtue” might be equally likely to be used by authors discussing family or business matters. Discussion of good behavior may span the contexts of nature, family history, legal cases, theology, mythology, etc. Across a large corpus of documents we may expect to see these themes arising in different combinations — both when different authors are writing within similar cultures, and when one author writes at different times in his or her career. Furthermore, writers write for different contexts and audiences: letters to friends or family or superiors, philosophical dialogues, public speeches, etc. Each of these contexts also changes the likelihood of the author selecting certain words, and the same word in different contexts may produce slight or major variations in meaning.

LDA topic modeling provides a method for automatically identifying topics within a set of documents. At the end of a training process:

(a) each topic is represented as a total probability distribution over all the words in the corpus — that is, every word is assigned a probability in every topic, and the sum of all the word probabilities within one topic is equal to one; and
(b) each document is represented as a total probability distribution over the topics — that is, every topic is assigned a probability in every document, and the sum of the topic probabilities within one document is likewise equal to one.

The model starts with random probabilities assigned to the word-topic and topic-document distributions. It is trained by a process of adjusting the word-topic and topic-document probability distributions. The word-topic and topic-document distributions are controlled by two parameters (technically “hyperparameters” or “priors”) that are set to ensure that there is sufficient variation in the probabilities assigned to the topics in the documents and to the words in the topics. The number of topics is chosen by the modeler. Our group typically trains multiple models with different numbers of topics, and we compare the different models to each other. For the present study we trained models with 20, 40, 60, 80, and 100 topics. In general, with too few topics, each topic becomes very general and hard to interpret. With too many topics, some of the topics are specialized on just a few documents, making them less useful for finding common themes. While there exist methods within computer science for estimating an optimally efficient number of topics for a given corpus, users of the models may prefer a coarse-grained scheme (fewer topics) for some purposes while other users may prefer a more fine-grained scheme (more topics) for other purposes.[32] Furthermore, working with multiple models simultaneously, fosters the kind of “interpretive pluralism” that characterizes humanities computing.[33]

The process by which we built these models using the InPhO topic-explorer package consists of four steps: initialization of the corpus object, preparation of the corpus by filtering words according to their frequency, training the corpus models, and launching the Topic Explorer’s Hypershelf interface.[34]

### Using the Topic Explorer & Notebooks

The Topic Explorer provides both a map-like visualization of the topic space (described further below) and a “Hypershelf” that allows users to experiment with the trained model to explore the corpus in any standard web browser. We call the latter interface a Hypershelf because although the browser initially presents documents from the corpus in a single linear order, it can be rearranged by the users to reflect their interests, and any document can be opened to view the full text. Thus, the Hypershelf initially provides a top level “distant reading”[35] view of the corpus, but allows the user to zoom down to the original text. This supports a two way interaction in which interpretation of the texts helps the user to interpret the topics in the model, while interpretation of the topics in the model can help the user to interpret the texts. (We provide an example of this interplay below.)

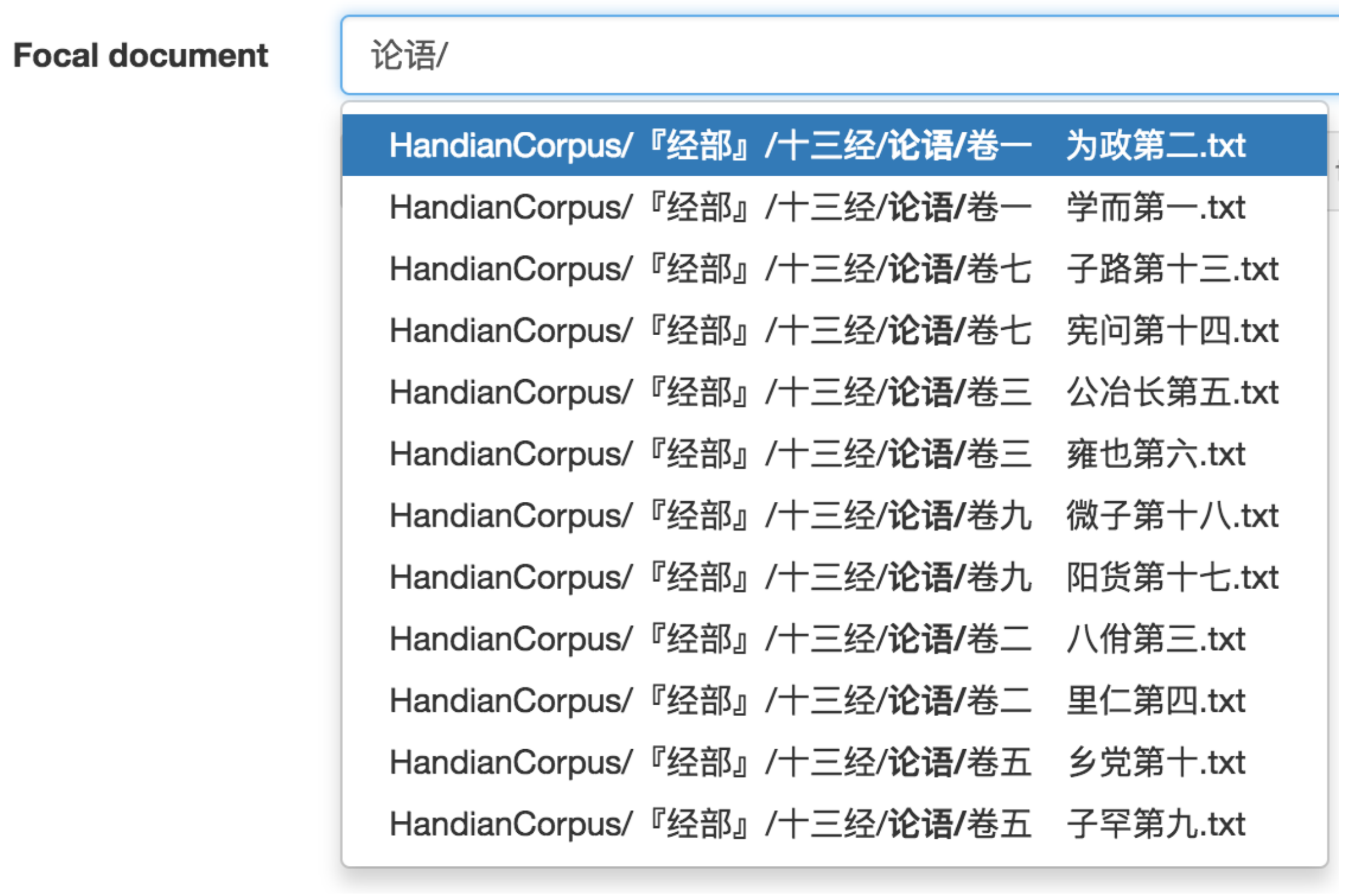

Figure 1. Autocompletion of document names.

The Hypershelf has two main modes: a document-centered view and a topic-centered view. Beginning with the document-centered view, the user may either select a document at random or use the search box to enter a few characters. These characters are automatically matched to the document labels, and the user can select a document from the drop-down list (Figure 1 shows initial options for 论语 — *Lúnyǔ*, the Analects). Once a document is selected, and a number of topics for the model is chosen, the browser window is filled with a row of multi-colored bars (Figure 2), each block of color corresponding to a topic. The top row represents the topics assigned to the document by the computer during the final training cycle, according to the key at the right. Hovering over any of the colored sections displays a list of the highest probability words for that topic (see Figure 3). It is important to remember, however, that every word is assigned a probability in every topic, so these first few words do not exhaust the context provided by the topic. Each subsequent row represents the topic distribution of another document from the corpus, scaled such that the length of the bar indicates similarity to the top document.[36]

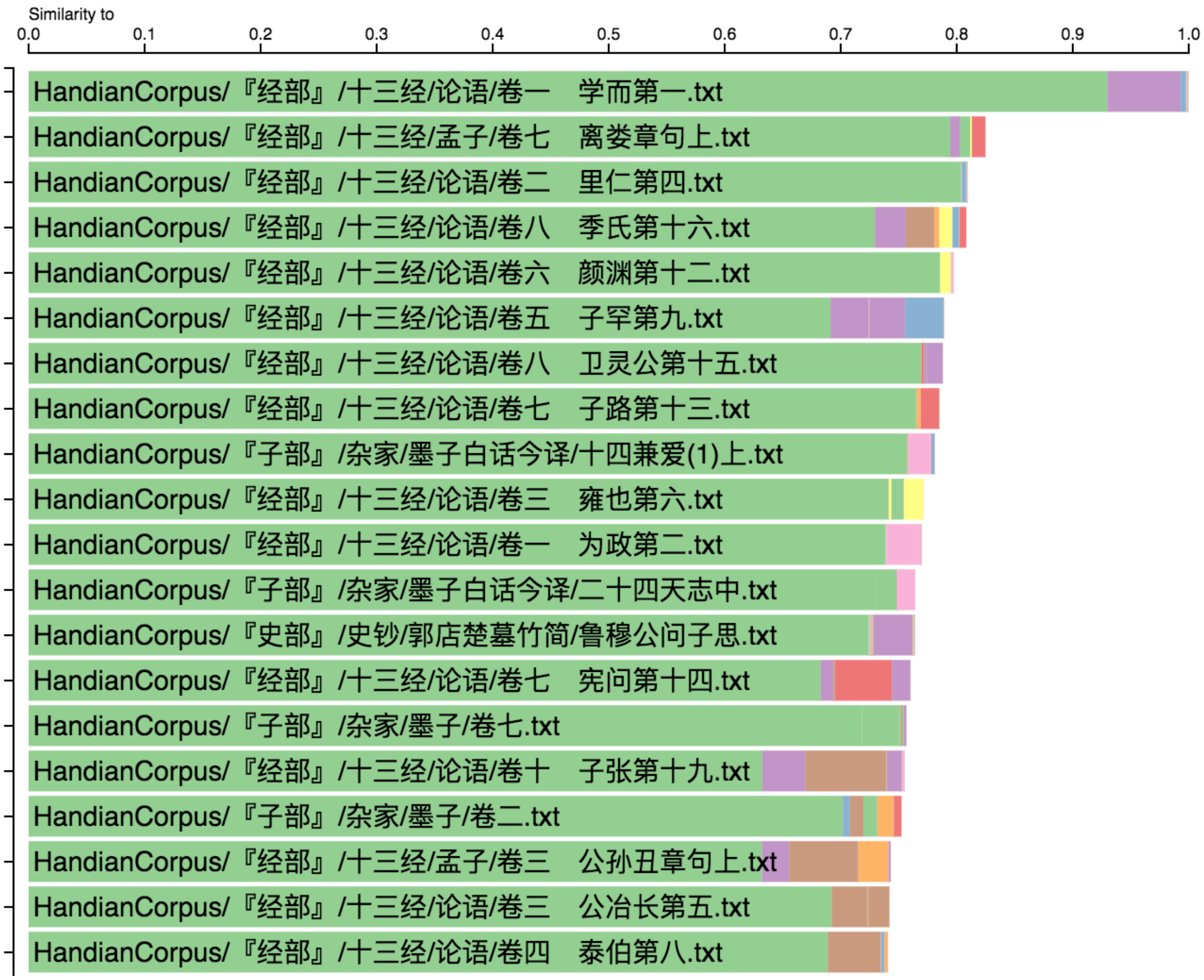

Figure 2. Similarity of documents to Book 1, Chapter 1 (卷一　学而第一) of the Analects (论语) in the 60-topic model. Each bar represents a document, and the colors represent the distribution of topics assigned to that document. (Different topics may be assigned to the same color.). The length of the bar indicates overall similarity to the document on the first row. See Figure 3 for additional details about the Topic Explorer display.

Initially, similarity between documents is shown with respect to the entire topic mixture associated with the focal document, but clicking the mouse on any of the topics re-sorts the list according to overall proportion of each document that the model assigned to the selected topic. This capacity of the HyperShelf allows users to rearrange the documents according to their interest in a particular topic (Figure 3).

From this point the user may click the “Top documents for Topic…” button below the key on the right (not shown in the Figure) to select the documents from the entire corpus that have been assigned the highest proportion of the selected topic. Alternatively, the user may choose to refocus on any of the other documents in the display by clicking on the arrow icon to the left of a row. (This icon appears when the mouse hovers nearby.) The user may read the full document by clicking on the "page" icon, which appears to the left of the arrow icon.

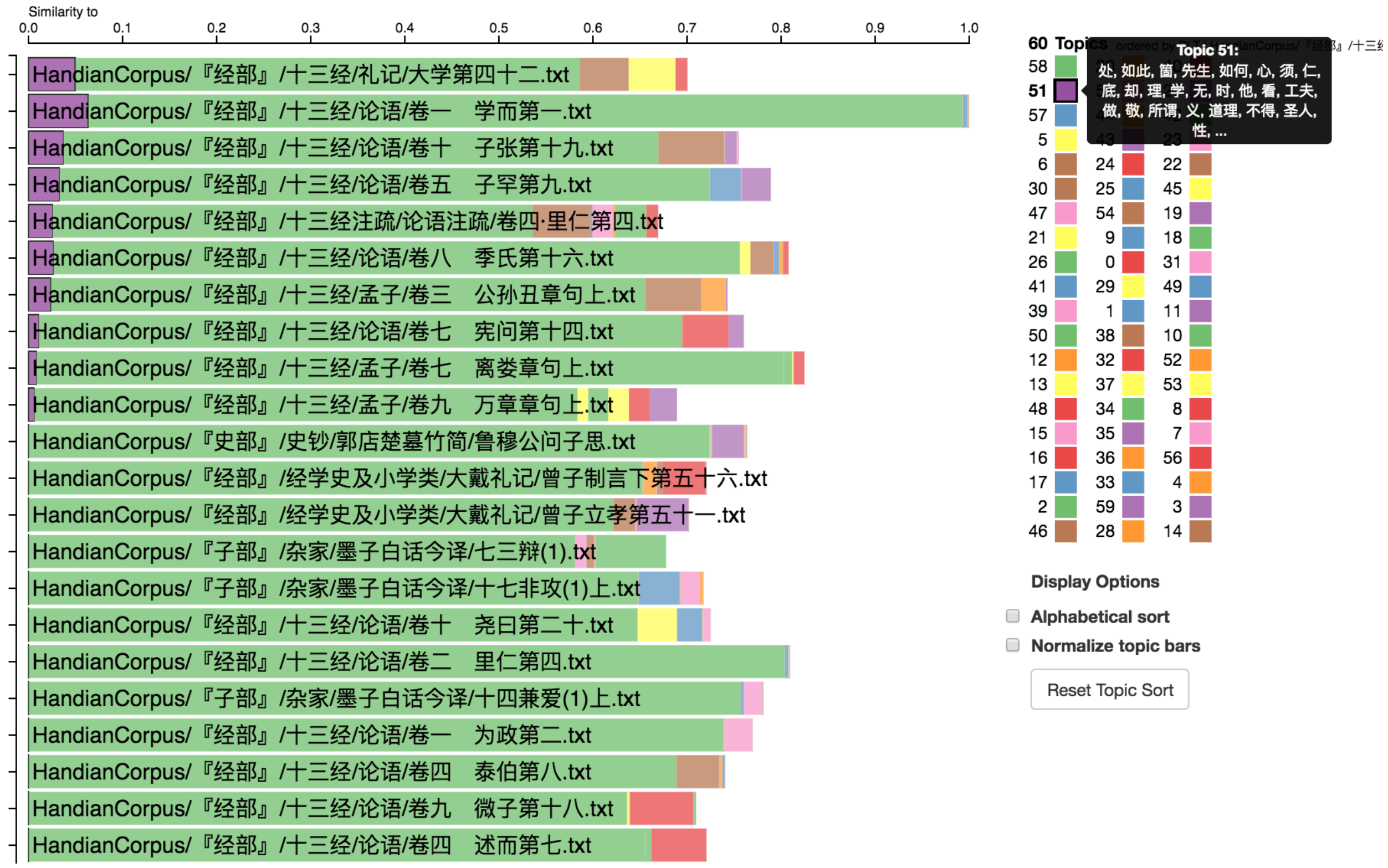

Figure 3. Highest probability words for each topic appear in the topic key at right when the cursor is positioned over the key, or over the corresponding topic in any of the document rows. Clicking in either location causes the Hypershelf to re-sort the list of results according to proportion of that topic assigned to each document. Here we show the reordering of results after selecting Topic 51 as the comparison dimension for documents most similar to Book 1 Chapter 1 (卷一　学而第一) of the Analects (论语) in the 60-topic model.

## Results

We successfully trained topic models on the corpus of over 18,000 classical Chinese documents and made them available to explore interactively online.[37] We believe our choice to do word segmentation rather than single character modeling is justified by the contribution that the two-character words make to the interpretability of the topics, as well as by our investigation of 阴阳 (yinyang) within the corpus, as described below.

Topic models for humanities computing cannot be evaluated against a "gold standard" of correct performance, because no such standard exists. Neither could such a standard exist if one takes seriously the idea that the process of interpretation at the core of the humanities applies to the models as much as the texts (see Discussion section below), and is as variable as the interests of the users themselves. Ultimately, a topic-modeling approach succeeds or fails according to the ability of users to use the models for their own purposes, be it self-education, pedagogy, exploratory research, or systematic analysis of the texts. In future work we intend to assess how users respond to the topic models, and to conduct more complete analyses of relationships among the texts using the models. Here we present an example of how a particular individual used the Topic Explorer modeling and visualization results for self-guided investigation and serendipitous discovery — a process we refer to as “guided serendipity”.[38]

Our subject, one of the Chinese coauthors of this paper, began this project with only a basic familiarity with ancient Chinese philosophy acquired from an undergraduate course. He decided initially to investigate the concept of yinyang (阴阳). Using the capacity of the Topic Explorer for topic-mediated term search, this term was queried in the 60 topic model.[39] Documents are retrieved according to their overall similarity to the topics selected by the term. The practical import is that because searches are topic-mediated, the documents retrieved need not contain the actual query term.

The first document identified in this way is from the 子部 (*zǐ bù*) section of the corpus, which contains writings of the philosophical schools. It is from the 术数 (shu shu, or divination) section of the corpus, specifically the 三命通会 (Sān Mìng Tōng Hui), an important book from the Ming dynasty. The specific chapter located in this way is 卷一·论支干源流 (On the Origin of the Chinese Sexagenary Cycle), describing the “ten Heavenly Stems” (yang) used in combination with “twelve Earthly Branches” (yin) as a calendar dating system.

A little further down the list of documents, in the 7th row is a chapter from the Confucianism subsection 儒家 (*rújiā* ). The chapter labeled ‘參兩篇第二’ from the volume labeled ‘张子正蒙’ in the corpus is part of the work *Zheng Meng* (正蒙), which is very significant within the Confucian tradition. It was written by Zhang Zai (1020-1077), an important thinker of the Song Dynasty from Shaanxi province. The chapter relates yinyang theory to the astronomical calendar and the classical theory of Five Phases: Wood, Fire, Earth, Metal, Water (also referring to Jupiter, Mars, Saturn, Venus and Mercury respectively) used to explain the laws governing speed and direction of planetary motion.

Pursuing the idea that the Five Phases Theory provides the backbone of a broad system of thought encompassing many areas, our subject re-sorted the Hypershelf by clicking the arrow to the left of the top row, to refocus on this document.[40] He then inspected the topics and identified topic 15 as seemingly most relevant to his interests. Clicking on that topic reorders the results according to the proportion of the topic allocated to each document in the list.

The top document identified in this way is also from the Confucian section of the Handian corpus, but in this case volume 12 from the book 春秋繁露 (*Chūn Qiū Fán Lù*, sometimes abbreviated as “Fan Lu”, and also known in English as the “Luxuriant Dew of the Spring and Autumn Annals) which relates the changing of the seasons to yinyang. By inspecting the titles of the documents near the top of the list, our subject noticed that many of them are from the *Chūn*

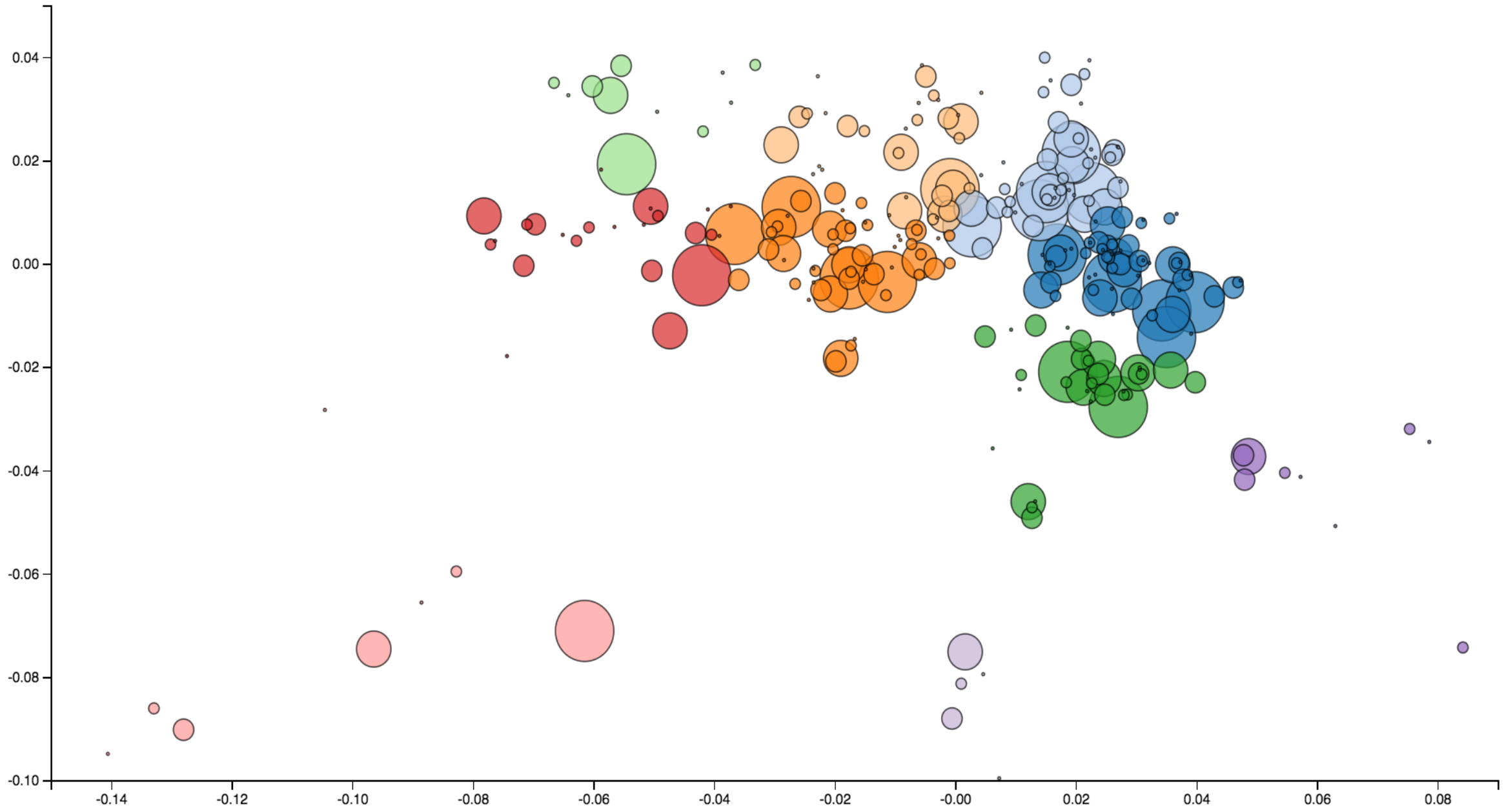

Figure 4. Topics from all five models arranged and clustered by the isomap algorithm. Circle size is inversely proportional to the number of topics in the model: largest circles representing topics from the 20-topic model, smallest from the 100-topic model. Overlapping circles of different sizes indicates congruence between topics from models at different levels. See main text for further details, and the next two figures for applications of the map to topics and terms

*Qiū Fán Lù*, from the *Zheng Meng,* and from a third important work titled '三命通会' (*Sān Mìng Tōng Huì*), which is a text about fortune telling and divination. This helped our subject to understand that the topic explorer could help him identify in which parts of Chinese culture the yinyang theory was prominent, namely Confucianism, Daoism, and traditional Chinese medicine.

For example, the document 卷二·论五行旺相休囚死并寄生十二宫 is part of the *Sān Mìng Tōng Huì* about the Five Phases theory, explaining the positive and negative relationships existing among Wood, Fire, Earth, Metal, and Water, and various ways in which those relationships may be in which these elements are related to human life, health, and death. Also present are numerous documents from the 医家 (Yījiā, or traditional medicine) section. Refocusing the topic explorer by clicking on  the arrow icon to the left of the document HandianCorpus/『子部』/医家/医学源流论/卷上·病同人异论.txt retrieves a large number of related medical texts.  In particular, the document from the 素问 (*Sù wèn*, or 'basic questions') section labeled 八正神明论篇第二十六 (Part 26 of the book Bā Zhèng Shénmíng Lùn) which is a very famous dialog between the mythologized "Yellow Emperor" Huángdì (黄帝) and his minister, Qí Bó (岐伯), supposed to have lived in the third millennium BCE.  They discuss acupuncture in the context of qi (气), a very important concept concerning life force or vital energy in traditional Chinese medicine, and they connect qi to yinyang.

Our coauthor reports that before using the Topic Explorer his concept of the Five Phases Theory was ambiguous, but in the interplay between topics and documents he learned many new details about the Five Phases Theory and its relationship to yinyang. For an expert, these interrelations may be well-known, but for a learner, the capacity to rapidly relate the concepts in this way serves a very valuable function. His understanding of the complexity of the concept of qi was also broadened, leading to a plan to work further on this concept in future work with the topic models. This example shows how one individual's understanding of the connectedness of concepts from traditional Chinese medicine, astronomy, and religion was deepened by interaction with both the high-level overviews provided by the topic model and the close reading of specific texts directed by following the models. Although just one case, we believe that this case is not unique: the Hypershelf interface of the Topic Explorer supports spontaneous exploration and guided serendipity, customized to the user's particular interests.

We turn now from the Hypershelf to an interactive visualization of the entire topic space which is also provided by the Topic Explorer software package. This visualization can be explored interactively at InPhO websites. Figure 4 shows a map and cluster analysis of the topics across all five models. The map is generated using the isomap procedure applied to the JSD measure.[41] Isomap is a technique for reducing a high dimensional space (in this case the probability space of the word distributions in the models) to fewer dimensions, in this case two. Such dimensionality reductions are useful for identifying principal components of the model structure. Whereas the standard MDS (multi-dimensional reduction) algorithms are linear, isomap detects non-linear structure in the data. The map allows one to assess the overall similarity of topics in the different models (20, 40, 60, 80, 100). The relationships among topics revealed by these figures are not strictly hierarchical; nevertheless, topics from the models with higher numbers of topics tend to cluster around topics from the models with smaller number of topics. Groups of topics are clustered and colored automatically according to an arbitrary choice of ten clusters. Although these clusters are very broad, some general themes emerge — for example, the dark green and dark purple clusters in the lower right are related to literature and poetry, the light blue region contains topics related to Confucianism, while the dark blue region below it spans topics related to traditional religions and traditional Chinese medicine. The light orange and dark orange regions cover different aspects of history; for example, topics related to military history are more prominent in the darker orange region. The dark red area corresponds to political and diplomatic topics while the pink cluster at the bottom left covers topics related to administration. The four topics colored light purple at the bottom center are heavily loaded with geographical terms — terms which are of course quite generally used in everything from poetry to military history and administration.

The interactive Topic Isomap supports exploration of the models in a variety of ways. Hovering the mouse over the elements of the map shows the highest probability words for the topics, and allows the user to click through to the Hypershelf, showing documents most similar to that topic. Alternatively, entering a word or words in the search box above the map adjusts the colors in the map to show the relative weight of the term across all the topics. More saturated colors indicate a higher probability of the term being generated by that topic. Figure 5 shows a comparison of the terms 孔子, Confucius (Figure 5a), and 佛, the Buddha (Figure 5b). Both are implicated in many of the topics, as indicated by the high number of topics receiving some color. However, the color map for 佛 shows a more concentrated set of topics belonging to the dark blue cluster. These topics, listed here, are all clearly related to Buddhism (X:Y indicates k=X, topic number Y):

20:1 师, 经, 僧, 佛, 时, 道, 寺, 无, 生, 法, 王, 如何, 真, 名, 处, …
40:19 师, 经, 僧, 佛, 时, 道, 无, 寺, 生, 法, 如何, 译, 王, 禅师, 处, …
60:5 经, 佛, 寺, 僧, 王, 法, 译, 时, 菩萨, 已, 沙门, 名, 身, 无, 善, …
80:37 佛, 寺, 经, 僧, 时, 王, 法, 无, 已, 身, 善, 菩萨, 名, 生, 受, …
100:96 经, 佛, 寺, 僧, 法, 译, 菩萨, 本, 沙门, 王, 释, 名, 塔, 录, 部, …

Clicking on any of the topics identified in this way takes the user to the Hypershelf with the top documents for that topic already loaded.

Figure 6 shows a similar comparison for the terms 气 (qi) and 阴阳 (yinyang). Here the distributions are quite similar, although the topics related to qi are more concentrated on the right side of the diagram whereas topics related to yinyang are distributed a bit more across central parts of the map. The relative confinement of topics related to qi corresponds to the fact that the Isomap algorithm has placed health and traditional medicine topics on the right side of the map in the dark blue cluster.

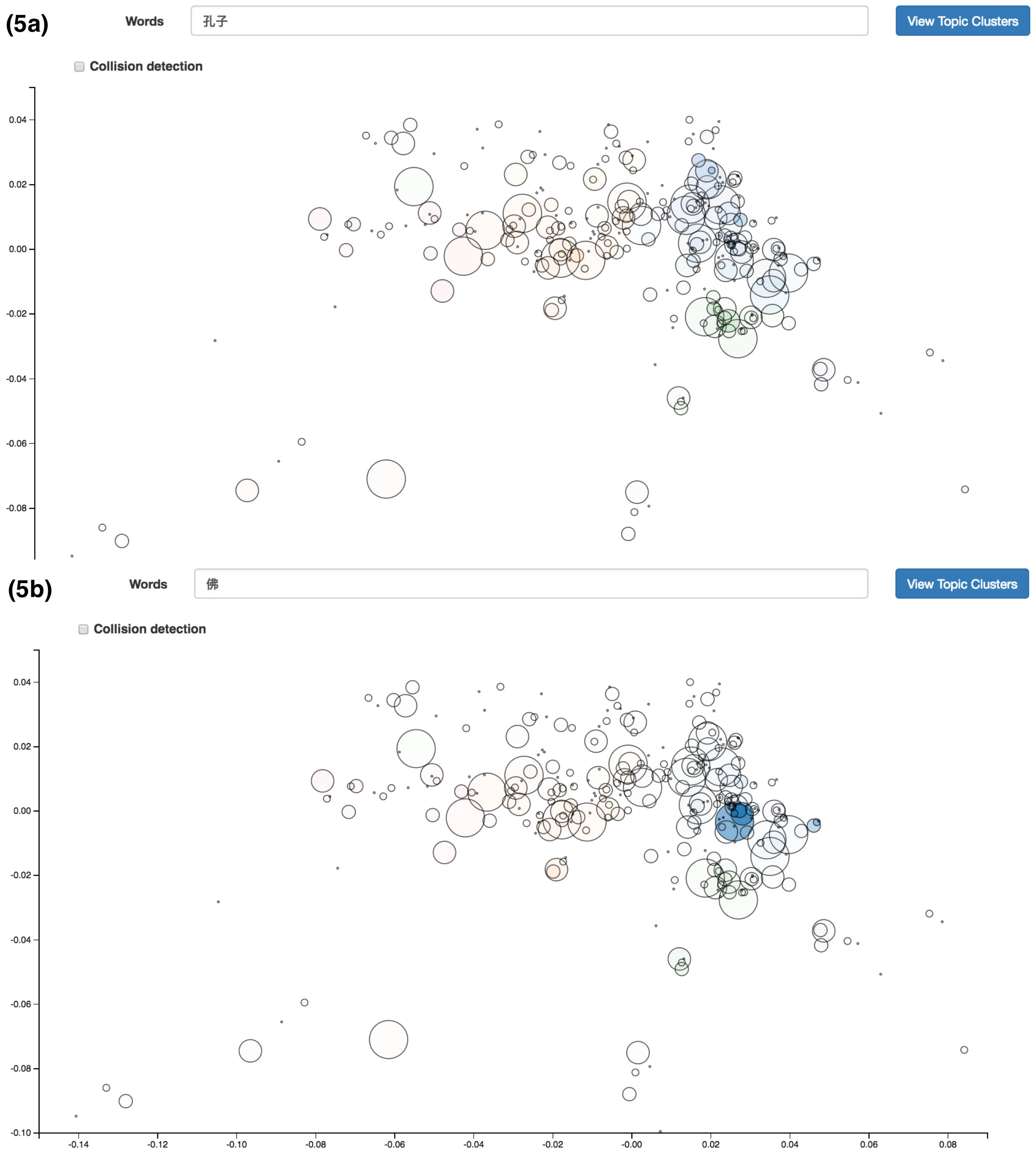

Figure 5. See figure 4 for explanation of underlying map. Here, colors are saturated according to relevance to word entered in the search box: (4a) 孔子, Confucius, and (4b) 佛, Buddha/Buddhism. Both are significant for many of the topics in the models, but 佛 selects a more specific set of topics, showing up as the saturated blue circles just to the right of the 0,0 origin of the map. See main text for further discussion.

Among the most central topics in the map (i.e., those closest to the 0,0 origin in the map) are these from right side of the dark orange cluster:

20:11 命, 官, 贼, 授, 兵, 尔, 巡抚, 营, 阿, 民, 大臣, 部, 明, 馀, 总督, …
40:23 命, 尔, 大臣, 授, 馀, 额, 阿, 营, 克, 总督, 部, 明, 巡抚, 布, 乾隆, …
60:18 命, 大臣, 馀, 总督, 授, 巡抚, 营, 议, 乾隆, 额, 署, 奏, 匪, 康熙, 由, …
80:40 馀, 命, 巡抚, 总督, 大臣, 议, 署, 匪, 由, 调, 乾隆, 奏, 免, 州县, 设, …
100:63 命, 大臣, 巡抚, 总督, 奏, 议, 学士, 谕, 乾隆, 署, 由, 授, 州县, 直, 调, …

These topics are highly loaded with terms related to government officials, but also contain some words related to criminality. The centrality of these topics may be seen as reflecting both the large number of government documents in the Handian corpus, and the central importance of the civil service in China for the preservation and transmission of classical Chinese culture and values. It is also worth noting that the 20:2 model (阙, 德, 臣, 无, 元, 圣, 表, 可, 命, 实, 天, 奉, 道, 文, 神, ...) is actually slightly closer to the center than 20:11, but it is grouped with the light blue cluster of topics. Visual inspection of the highest probability terms suggests that 20:11than 20:2 is more aligned with the other topics listed above, and helps give some confidence in the clustering technique. It is important to keep in mind, however, that the isomap plot is generated using the full term distribution, not just the first 15 terms shown here. A complete assessment of the topics and their related documents would go beyond simple inspection of the top terms.

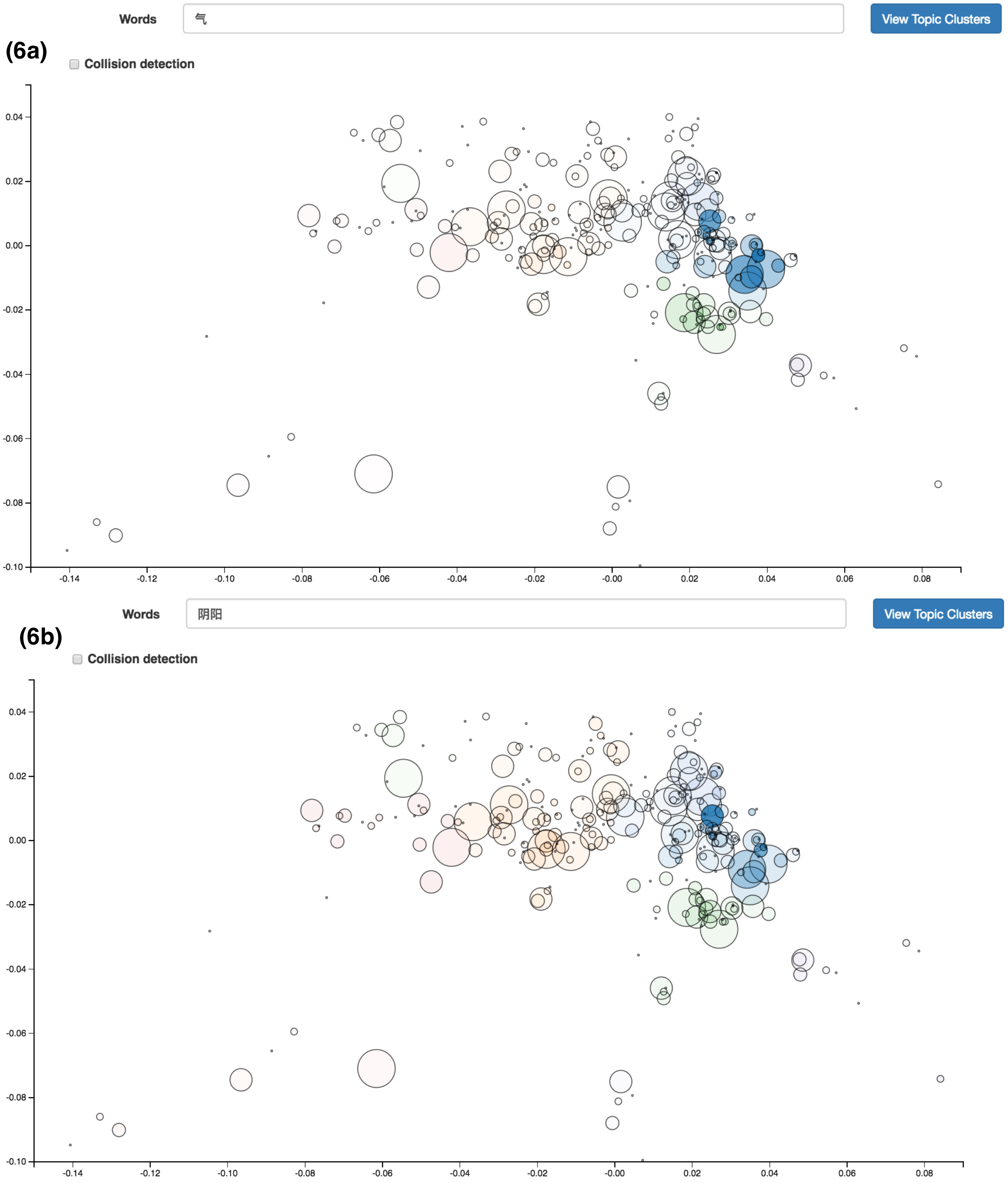

Figure 6. See Figures 4 and 5 for explanation of layout and coloring scheme. Here we compare distribution of topics for two terms：(6a) 气, qi, and (6b) 阴阳, yinyang. The distribution of these topics in the corpus is rather similar, but 气 is slightly more concentrated on the right side of the map, where topics related to health and medicine are clustered.

## Discussion

LDA topic models are not themselves interpretations of the documents — indeed they stand in need of interpretation themselves[42] — but they may assist scholars in exploring and interpreting large collections of materials. Ultimately there is no substitute for reading the documents, but the Topic Explorer interface, through its Hypershelf and Topic Isomap components, can guide scholars and learners alike to documents that they might not have otherwise encountered or thought to look for, resulting in a particularly productive form of guided serendipity.

Topic models are interesting to think about from the perspective of theories of meaning. While they do not capture exact meanings — "John loves Mary" and "Mary loves John" are viewed as identical statements under the "bag of words" assumption — they are quite successful at capturing something like the general gist or context of the words being used. Scholars of Chinese literature have emphasized the high degree of context sensitivity for the meanings of words in the Chinese language,[43] but a strength of the topic modeling approach is that the same word is placed in multiple contexts, helping with the process of disambiguation. At the same time, because the models have a solid grounding in information theory, the use of metric measures such as the Jensen-Shannon distance is feasible for many applications. This provides new forms of evidence for humanistic discussions.

Although the corpus we used may be missing some potentially important documents, it is large enough that the topic models we derived from this corpus prove to be adequate for various purposes. Improved curation of the corpus nevertheless remains an important goal of our group for the future, and will be reflected in future iterations of the Handian Topic Explorer mirror site. Future work will allow us to address questions about topical relationships among the documents in the Handian corpus and about historical and geographical shifts in the topic distributions as represented in the corpus model, and ultimately to analyze the behavior of individual authors.

Finally, and more speculatively, philosophers of mind and cognitive science have sometimes been tempted by the idea that meaning or semantic content assignment is a kind of measurement process rather than the assertion of a relationship to a determinate abstract proposition.[44] Computer scientists have started to provide the means to convert this idea into quantitative models[45] to which measures such as Jensen-Shannon distance can be applied. Thus, the Digital Humanities are poised to have a significant impact on philosophical and practical discussions of the nature of meaning.

## Conclusions

Topic models present a powerful new tool for computer-assisted interpretation in the humanities. We have described some particular issues faced for using topic models with ancient Chinese texts, and we have detailed the process of training LDA topic models on the Handian corpus of over 18,000 classical Chinese texts using the InPhO Topic Explorer. The results of these efforts and the software we have developed have been made publicly available via the Hypershelf interface at mirror sites at Xi'an Jiaotong University and Indiana University. This interface allows users to visualize the results of the modeling process. We have provided some preliminary description and analysis of the topics discovered by the algorithms using the more advanced notebook features of the Topic Explorer. These preliminary investigations have revealed some interrelationships among Confucian, Taoist and Buddhist themes, and the penetration of these themes in many aspects of traditional Chinese culture, from medicine to government. By following the threads among specific texts, guided by these topic models, scholars may exploit these new tools to enrich their understanding and interpretation of China's rich cultural heritage.

## Acknowledgements

The software described in this paper was originally developed at Indiana University with generous support from the National Endowment for the Humanities and IU's Office of the Vice President for Research. Its extension to ancient Chinese owes much to IU's support and to the research funding provided by the College of Humanities and Social Sciences, the office of Dean Yanjie Bian and the Philosophy Department at Xi'an Jiaotong University. The authors of this paper, we would like to thank Xiaoliang Wang and Wenjing Yuan at XJTU for their initial guidance concerning ancient Chinese philosophy. We also acknowledge the prior programming efforts of Robert Rose, Doori Lee, Jessie Pusateri, and Adithya Nagaraj-Tirumale at Indiana University. We are grateful to Henry Rosemont, Jr., for comments on the manuscript. All errors of interpretation remain our own.

## Notes

[1] Paul Tasman, "Literary data processing," *IBM Journal of Research and Development* 1, no. 3 (1957), 249-256. Stephen M. Parrish, *A Concordance to the Poems of Matthew Arnold*. *Ithaca, NY: Cornell University Press, 1959). Robert J. Glickman and Gerrit J. Staalman, *Manual for the Printing of Literary Texts and Concordances by Computer* (University of Toronto Press, 1966). John B. Smith, "Image and imagery in Joyce's Portrait: A computer-assisted analysis," in Stanley Weintraub (Ed.) *Directions in Literary Criticism: Contemporary Approaches to Literature*. (University Park, PA: Pennsylvania State University Press, 1973). John B. Smith, "Computer criticism," *Style* 12, no. 4 (1978), 326-356. For a historical overview, see Geoffrey Rockwell and Stéfan Sinclair, *Hermeneutica: Computer-Assisted Interpretation in the Humanities* (Cambridge, MA: The MIT Press, 2016), chapter 3.

[2] Rockwell & Sinclair, *ibid.*

[3] http://hathitrust.org/.

[4] Heather Christenson, "HathiTrust: A Research Library at Web Scale," *Library Resources and Technical Services* 55 (2011), 93-102.

[5] https://www.hathitrust.org/statistics_info

[6] Stacy T. Kowalczyk, Yiming Sun, Zong Peng, Beth Plale, Aaron Todd, Loretta Auvil, Craig Willis, Jiaan Zeng, Milinda Pathirage, Samitha Liyanage, Guangchen Ruan and Stephen J. Downie, "Big Data at Scale for Digital Humanities: An Architecture for the HathiTrust Research Center," in Wen-Chen Hu and Naima Kaabouch (Eds.) *Big Data Management, Technologies, and Applications* (Hershey, PA: IGI Global, 2014), 345-369.

[7] Such projects include: Barbara B. Oberg and J. Jefferson Looney (Eds.) *The Papers of Thomas Jefferson Digital Edition (*Charlottesville, VA: University of Virginia Press, 2009—); the correspondence of Charles Darwin (www.darwinproject.ac.uk); the Folger Shakespeare Library (http://www.folger.edu/folger-digital-texts); a corpus of Chinese historical records by Hui Dong, Lei Xu, Fei Wang and Siwei Yu, "Study on Semantic Analysis System (III) —Implementation of Chinese Historical Records Semantic Analysis System," *Journal of the China Society for Scientific and Technical Information* 33, no. 2 (2014), 204-214; and too many other projects to mention.

[8] http://gutenberg.org/

[9] Susan T. Dumais, "Latent Semantic Analysis," *Annual Review of Information Science and Technology* Volume 38 (2004), chapter 4, 189-230. David M. Blei, Andrew Y. Ng, and Michael I. Jordan, "Latent Dirichlet Allocation," *Journal of Machine Learning Research*, 3 (2003), 993–1022. Tomas Mikolov, Kai Chen, Greg S. Corrado and Jeffrey Dean, "Efficient Estimation of Word Representations in Vector Space," ArXiv e-prints (2013), https://arxiv.org/abs/1301.3781 (retrieved Dec 21, 2016).

[10] Sarah Klingenstein, Tim Hitchcock and Simon DeDeo, "The civilizing process in London's Old Bailey," *Proceedings of the National Academy of Sciences* 111, no. 26 (2014), 9419-9424. Andrew Goldstone and Ted Underwood, "The Quiet Transformations of Literary Studies: What Thirteen Thousand Scholars Could Tell Us," *New Literary History* 45 (2014), 359-384. Ted Underwood and Jordan Sellers, "The Longue Durée of Literary Prestige," *Modern Language Quarterly* 77, no. 3 (2016), 321-344.

[11] Lauren Klein and Jacob Eisenstein, "Reading Thomas Jefferson with TopicViz : Towards a Thematic Method for Exploring Large Cultural Archives," *Scholarly and Research Communication* 4 (2013), 1–12. Jaimie Murdock, Colin Allen and Simon DeDeo, "Exploration and Exploitation of Victorian Science in Darwin's Reading Notebooks," *Cognition* 159 (2017), 117–126.

[12] Jian Ouyang, "Visual Analysis and Exploration of Ancient Texts for Digital Humanities Research," *Journal of Library Science in China* 42, no. 222 (2016), 66-80.

[13]Zixuan Huang and Jinsong Yu, "Topic Model based SongCi Corpus Construction and Research on Computer aided SongCi Writing," *International Journal of Knowledge and Language Processing* 3, no. 2 (2012), 1-19.

[14] http://ctext.org/.

[15] Ryan Nichols, Kristoffer L. Nielbo and Uffe Bergeton, "Topic Modeling the Ancient Chinese Corpus: The Textual Contexts of High and Low Gods in Chinese Thought." Conference presentation at Cultural Evolution of Religion Research Consortium, Montreal, Canada, May 2015. Ryan Nichols, Edward Slingerland, Kristoffer Nielbo, Uffe Bergeton, Carson Logan and Scott Kleinbahn, "Modeling the contested relationship between *Analects*, *Mencius*, and *Xunzi*: Preliminary evidence from a Machine-Learning Approach," submitted. Slingerland, Nichols, Nielbo and Logan, "The Distant Reading of Religious Texts: A "Big Data" Approach to Mind-Body Concepts in Early China," submitted.

[16] Rockwell & Sinclair *op cit.* David M. Blei, "Probabilistic Topic Models," *Communications of the ACM* 55 (2012), 77–84. Ted Underwood, "Theorizing Research Practices We Forgot to Theorize Twenty Years Ago," *Representations* 127 no. 1 (2014), 64-72.

[17] Henry Rosemont, Jr. "Translating and Interpreting Chinese Philosophy," *The Stanford Encyclopedia of Philosophy (Summer 2016 Edition)*, Edward N. Zalta (ed.), URL http://plato.stanford.edu/archives/sum2016/entries/chinese-translate-interpret.

[18] Qi Zhao Q, Zengchang Qin and Tao Wan, "Topic Modeling of Chinese Language Using Character-Word Relations," in: Bao-Liang Lu, Liqing Zhang and James Kwok (Eds.), *Proceedings of the 18th International Conference on Neural Information Processing,* ICONIP 2011. *Lecture Notes in Computer Science*, vol 7064. (Berlin, Heidelberg: Springer, 2011).

[19] Rosemont *op. cit.*

[20] Jaimie Murdock and Colin Allen, "Visualization Techniques for Topic Model Checking," *Proceedings of the Association for the Advancement of Artificial Intelligence (AAAI).* (Austin, TX: AAAI Press, 2015), http://www.aaai.org/ocs/index.php/AAAI/AAAI15/paper/view/10007.

[21] http://online.eastview.com/projects/skqs/

[22] http://www.zdic.net/aboutus/ (retrieved November 1, 2016).

[23] Nichols et al. 2015 and submitted, Slingerland et al. submitted.

[24] http://inphodata.cogs.indiaa.edu/handian/ and http://inpho.xjtu.edu.cn/handian/.

[25] The 2nd-level sections of the Handian corpus are listed in appendix 1 of the supplemental materials available at <http:URL>.

[26] After training the topic models, we discovered, with the help of the models, that two folders of English translations of the Analects (20 files) and Lao Tze (84 files) had slipped through our net.

We decided not to remove these and retrain the models because the presence of these files has a minimal effect on the overall results.

[27] Zhao, Qin & Wang *op. cit.* Ouyang *op. cit.*

[28] Corpus cleaning methods are detailed in appendix 2 of supplemental materials. The ancient words dictionary is available at https://github.com/inpho/topic-explorer/blob/master/topicexplorer/lib/ancient%20words.dic and embedded within the Topic Explorer using the "och" (old Chinese) extensions.

[29] Gerard Salton, Andrew Wong and Chungshu Yang, "A Vector Space Model for Automatic Indexing," *Information Retrieval and Language Processing* 18 no. 11 (1975), 613-620. Christopher Fox, "A stop list for general text," *SIGIR Forum* 24 nos. 1-2 (1989), 19-21. Steven Bird, Ewan Loper and Edward Klein, *Natural Language Processing with Python*, (Sebastopol, CA, O'Reilly Media Inc., 2009).

[30] Supplemental materials appendix 3 shows the 20 most common words, and their frequencies, and describes the process by which we developed our list of 187 words for ancient Chinese in Appendix 3. For the stop word list used by Slingerland et al. see their "Modeling the contested relationship", submitted.

[31] Blei, "Probablistic Topic Models."

[32] Hanna M. Wallach, Iain Murray, Ruslan Salakhutdinov and David Mimno, "Evaluation Methods for Topic Models," in *Proceedings of the 26th Annual International Conference on Machine Learning, ICML '09* (New York, NY, USA: ACM, 2009), pp. 1105–12. Margaret Roberts, Brandon Stewart and Dustin Tingley, "Navigating the Local Modes of Big Data: The Case of Topic Models," in *Data Analytics in Social Science, Government, and Industry* (New York: Cambridge University Press, 2015). Jonathan Chang, Sean Gerrish, Chong Wang, Jordan L Boyd-Graber and David M Blei, "Reading Tea Leaves: How Humans Interpret Topic Models," in *Advances in Neural Information Processing Systems*, 2009, pp. 288–96.

[33] Rockwell and Sinclair *Hermeneutica*.

[34] Details of the modeling process provided in supplemental appendix 4. The topics discovered by training on the Handian corpus are represented in appendix 5, which shows the 15 highest probability words from each topic in the 20, 40, 60, 80, and 100 topic models.

[35] Franco Moretti, *Distant Reading* (New York: Verso, 2013).

[36] Jensen-Shannon Distance is due to Dominik M. Endres and Johannes E. Schindelin, "A New Metric for Probability Distributions," *IEEE Transactions on Information Theory* 49, no. 7 (2003), 1859-1861. It is based on Kullback-Leibler divergence introduced by Solomon Kullback and Richard A. Leibler, "On information and sufficiency," *Annals of Mathematical Statistics* 22, no. 1 (1951), 79–86. See supplemental appendix 6 for additional information.

[37] The models may be explored interactively at http://inphodata.cogs.indiana.edu/handian/ or at http://inpho.xjtu.edu.cn/handian/. See also supplemental appendix 5.

[38] Cameron Buckner, Matthias Niepert and Colin Allen, "From encyclopedia to ontology: toward dynamic representation of the discipline of philosophy," *Synthese* 182 (2011), 205-233.

[39] http://inphodata.cogs.indiana.edu/handian/60/?q=阴阳.

[40] Equivalent to http://inphodata.cogs.indiana.edu/handian/60/?doc=HandianCorpus%2F『子部』%2F 儒家%2F 张子正蒙%2F 參兩篇第二.txt.
[41] Joshua B. Tennenbaum, Vin de Silva and John C. Langford, "A Global Geometric Framework for Nonlinear Dimensionality Reduction," *Science* 290, no. 5500 (2000), 2319-2323. We applied the isomap algorithmhm to the JSD measure.
[42] Blei, "Probabilistic Topic Models"; Rockwell and Sinclair *Hermeneutica*; Underwood, "Theorizing Research Practices."
[43] Rosemont, "Translating and Interpreting Chinese Philosophy."
[44] For example, Paul M. Churchland, *Scientific Realism and the Plasticity of Mind* (Cambridge: Cambridge University Press, 1979), and Robert Matthews, *Measure of Mind* (New York: Oxford University Press, 2007).
[45] For a review see Michael N. Jones, Thomas M Gruenenfelder and Gabriel Recchia, "In Defense of Spatial Models of Lexical Semantics," in Laura Carlson, Christoph Hölscher and Thomas Shipley (Eds.), *Proceedings of the 33rd Annual Conference of the Cognitive Science Society* (Austin, TX: Cognitive Science Society, 2011), 3444–3449.

# Topic Modeling the *Hàn diăn* Ancient Classics (汉典古籍)

Colin ALLEN[1,2], Hongliang LUO[3], Jaimie MURDOCK[2], Jianghuai PU[1], Xiaohong WANG[1], Yanjie ZHAI[1], Kun ZHAO[3]

[1] Department of Philosophy, School of Humanities and Social Sciences, Xi'an Jiaotong University, Shaanxi, China
[2] Cognitive Science Program, Indiana University, Bloomington, Indiana, USA
[3] Institute of Computer Software and Theory, School of Electronic and Information Engineering, Xi'an Jiaotong University, Shaanxi, China

Authors listed alphabetically

**SUPPLEMENTAL MATERIAL**

**Abstract:** Humanities scholars have become increasingly interested in exploiting computational methods to assist in the development of new insights and interpretations of culturally significant materials. Ancient Chinese texts present an area of enormous challenge and opportunity. In this paper we describe a collaborative effort between Indiana University and Xi'an Jiaotong University to support exploration and interpretation of a digital corpus of over 18,000 ancient Chinese documents, which we refer to as the "Handian" ancient classics corpus (汉典古籍 or *Hàn diăn gŭ jí*, i.e, the "Han canon" or "Chinese classics"). It contains classics of ancient Chinese philosophy, documents of historical and biographical significance, and literary works. We begin by describing the Digital Humanities context of this joint project, and the advances in humanities computing that made this project feasible. We describe the corpus and introduce our application of probabilistic topic modeling to this corpus, with attention to the particular challenges posed by modeling ancient Chinese documents. We give a specific example of how the software we have developed can be used to aid discovery and interpretation of themes in the corpus. We outline more advanced forms of computer-aided interpretation that are also made possible by the programming interface provided by our system, and the general implications of these methods for understanding the nature of meaning in these texts.

**Appendix 1:** Level 2 subsections of the Handian classics corpus

| | | | |
|---|---|---|---|
| 经部（*Jīng Bù*）Classics | : | 十三经 | the thirteen classics |
| | | 十三经注疏 | thirteen classics annotations |
| | | 经学史及小学类 | history of classical studies and traditional philology |
| 史部（*Shǐ Bù*）Histories | : | 正史 | official dynastic histories |
| | | 编年 | annals and chronicles |
| | | 纪事本末 | complete accounts of major historical events |
| | | 别杂史等 | alternative and miscellaneous histories |
| | | 史评 | historical commentaries |
| | | 传记 | biographies |
| | | 载记 | regional histories |
| | | 地理 | geography |
| | | 职官 | state functions and powers |
| | | 政书 | Book of Zheng (politics and policy) |
| 子部（*Zǐ Bù*）Masters | : | 儒家 | Confucianism |
| | | 释家 | Buddhism |
| | | 道家 | Taoism |
| | | 法家 | legalism |
| | | 兵家 | School of the Military |
| | | 农家 | School of Agronomy |
| | | 杂家 | miscellaneous schools |
| | | 术数 | numerology |
| | | 医家 | traditional medicine |
| | | 科技 | science and technology |
| | | 艺术 | art |
| | | 书画 | calligraphy and painting |
| | | 谱录 | miscellaneous treatises |
| | | 类书 | reference works |
| | | 蒙学 | basic education |
| 集部（*Jí Bù*）Belle-lettres | : | 总集 | major anthologies |
| | | 别集 | secondary anthologies |
| | | 楚辞 | Songs of Chu / Songs of the South |
| | | 词 | lyrics |
| | | 诗文评 | reviews of poetry and prose |
| | | 曲 | Yuan dynasty opera |
| | | 小说 | fiction |

**Appendix 2:** Corpus cleanup

1. 38,182 HTML files were downloaded from gj.zdic.net.
2. These files were processed to extract the primary text within an html container marked by zdic.net with their div="snr2" container.
3. All HTML tags within this text were removed.
4. Files were converted, if necessary, to use the unicode "utf-8" specification and saved with the name of the document and a .txt extension.
5. All traditional Chinese characters were converted to simplified Chinese using the Python hanziconv package (https://pypi.python.org/pypi/hanziconv, version 0.3.2). Fortunately, the mapping between the two character sets is 1-1.
6. We removed a subfolder of the 集部 section of the corpus labeled 小说 that contained 18,049 files corresponding to modern fiction in order to focus on the classical corpus. This left 20,133 files.
7. We removed empty files leaving 19,865 files.
8. Each file was tested for the presence of modern Chinese words. 1,459 files thus identified were manually inspected to see whether they should be discarded, or edited by hand to extract ancient text only. 412 of these were determined to be important and kept in the corpus after being manually edited to remove the modern Chinese sections. Thus we eliminated 1,047 files, leaving 18,818 files.

**Appendix 3:** Highest frequency words

*Top 20 words in the unfiltered corpus, with frequencies*

| | | | | | | | |
|---|---|---|---|---|---|---|---|
| 之 | 1275107 | 其 | 566797 | 不 | 326500 | 与 | 230072 |
| 以 | 759017 | 为 | 565127 | 是 | 318320 | 在 | 225898 |
| 也 | 648940 | 者 | 560030 | 则 | 278791 | 州 | 218932 |
| 有 | 608156 | 曰 | 531502 | 自 | 237334 | 一 | 217999 |
| 而 | 587775 | 于 | 502754 | 云 | 232693 | 至 | 216245 |

We developed a stopword list that includes 187 words, including 19 of these 20 most frequent terms (all but 州, *zhōu*, state or prefecture). The list was developed using a mixed strategy of (a) checking word frequencies; (b) checking inverse document frequencies (i.e., the logarithm of the ratio of the total number of documents to the number of documents containing the word — a low score means the word is highly distributed throughout the corpus with a score of zero meaning that the word is present in every file); and (c) a cycle of training and retraining topic models, inspecting the topics, and adding terms to the stop list if they occurred in a high proportion of the topics and were judged uninformative about the topic.

The criteria we considered in deciding whether to place a term on the stop list were: 1. Does the term sometimes have a culturally/philosophically significant meaning? 2. Where it appears among the highest probability words for a topic, is it most likely to be interpreted with that significant meaning? 3. If it were removed from those topics, would the topics be less interpretable? The resulting list of 187 stop words is available at http://inphodata.cogs.indiana.edu/handian/cn_stop.txt. The top 20 words with their frequencies after the stopword list was applied are shown below.

*Top 20 words in the filtered corpus with frequencies*

| | | | | | | | |
|---|---|---|---|---|---|---|---|
| 州 | 218932 | 官 | 160995 | 书 | 107519 | 卷 | 102601 |
| 无 | 200605 | 令 | 146193 | 县 | 107071 | 已 | 102028 |
| 王 | 172465 | 行 | 132642 | 文 | 104966 | 命 | 101657 |
| 臣 | 164252 | 将 | 116283 | 兵 | 104701 | 道 | 101096 |
| 时 | 163895 | 帝 | 109950 | 诏 | 103798 | 初 | 101023 |

**Appendix 4:** The Topic Explorer modeling process

The InPhO Topic Explorer software can be freely downloaded at https://github.com/inpho/topic-explorer. User and developer versions of the software are installed in a few easy steps as explained in the site's Readme file which is also displayed at the URL above. There are four main phases to generating a functioning Topic Explorer instance. All commands are run in a shell or terminal window, depending on the operating system used (linux, mac os x, and Windows are supported).

1. 'topicexplorer init <*corpus_location*>' requires the user to specify the location of the corpus. The notation <…> indicates an item to be supplied by the user, in this case *corpus_location* should identify a directory or folder containing all the files to be modeled. By default, the software will suggest the name of the directory corresponding to the *corpus_location* as the *corpus_name.* The init step prompts the user to choose a project name (by default the same as corpus_location, and it creates a corpus file which records the count of every word in each document. The corpus file is placed in a folder called 'models'. Also, it creates a configuration ".ini" file using the project name, i.e. *corpus_name.ini* for whatever corpus_name the user supplied. For ancient Chinese word segmentation we run the init step with the special tokenizer flag 'ltc', which uses pymmseg-cpp' — a Python interface to a high performance implementation of mmseg in the Ruby language. We supplemented mmseg with a list of 3,636 multi-character words in ancient Chinese that we extracted from https://zh.wiktionary.org/zh/User:Wihwang/古漢語常用字字典. The resulting list is available at https://github.com/inpho/topic-explorer/blob/master/topicexplorer/lib/ancient%20words.dic. By default, the software filters out all words occurring 5 or fewer times in the corpus, although this behavior can be controlled with the --freq flag. Summing up:
    topicexplorer init <*corpus_location*> --tokenizer ltc

2. 'topicexplorer prep <*config_file*>' allows the user to apply a stopword list and interactively set low and high thresholds for excluding the most common and the most rare words. After the high and low thresholds are set, an updated corpus file is placed in the models directory. Summary:
    topicexplorer prep <*config_file*> --stopword_file <file_name>

3. 'vsm train <*config_file*>' allows the user to select interactively the number of topics in each model, and the number of training cycles. Training in the InPhO Topic Explorer proceeds via a version of the widely-used Gibbs sampling method. For this present study we trained models with 20, 40, 60, 80 and 100 topics for 1000 training cycles. Summary:
    topicexplorer train <*config_file*> -k 20 40 60 80 100 --iter 1000

4a. 'launch' starts an instance of the interactive Topic Explorer Hypershelf in the web browser:
    topicexplorer launch <*config_file*> --fulltext
The optional argument flag '--fulltext' makes the full texts accessible from within the Hypershelf.

4b. 'notebook' starts up an interactive Python notebook server in Jupyter:
    topicexplorer notebook <*config_file*>
We do not provide a detailed explanation of the use of notebooks in this paper. However, running the topicexplorer notebook command produces a tutorial notebook for the corpus which provides a basis for further exploration.

## **Appendix 5:** The Handian topics

The models described in this work were trained in December 2016 by the first author following the steps outlined in Appendix 4 on a MacBook Pro. The resulting model files are archived at <<URI>>. Here we show the 15 highest probability words for each topic in the five models trained from the Handian corpus archived at <<URI>>. Topic labels (index numbers) are arbitrarily assigned by the computer, and have no intrinsic significance. Future iterations of the inpho/Handian web site may not correspond exactly to the topics shown here. This is because repetition of the modeling process with the same corpus and parameters typically produces slightly different results due to stochastic aspects of the modeling process. (See Murdock & Allen 2015 for discussion.) Additional variation will result as we upgrade the corpus and make other adjustments to model parameters.

***20 topic model, showing 15 highest probability words in each topic***

| | |
|---|---|
| Topic 0 | 為, 於, 無, 國, 軍, 後, 則, 書, 與, 長, 謂, 時, 禮, 諸, 東 |
| Topic 1 | 师, 经, 僧, 佛, 时, 道, 寺, 无, 生, 法, 王, 如何, 真, 名, 处 |
| Topic 2 | 阙, 德, 臣, 无, 元, 圣, 表, 可, 命, 实, 天, 奉, 道, 文, 神 |
| Topic 3 | 服, 气, 治, 病, 热, 汤, 脉, 血, 生, 寒, 升, 丸, 阴, 钱, 散 |
| Topic 4 | 星, 犯, 月, 日, 朔, 天, 占, 岁, 火, 太白, 雨, 木, 赤, 荧惑, 辰 |
| Topic 5 | 将军, 刺史, 王, 帝, 州, 令, 尚书, 郡, 太守, 封, 魏, 迁, 时, 初, 诏 |
| Topic 6 | 臣, 陛下, 无, 议, 时, 奏, 已, 官, 可, 书, 论, 天下, 不可, 以为, 罪 |
| Topic 7 | 王, 侯, 汉, 秦, 帝, 国, 立, 令, 魏, 封, 赵, 臣, 楚, 时, 将 |
| Topic 8 | 卷, 花, 春, 归, 无, 时, 客, 山, 空, 秋, 日, 风, 尽, 酒, 送 |
| Topic 9 | 诗, 赋, 歌, 纪, 玉, 山, 游, 飞, 集, 风, 时, 无, 类聚, 本, 御 |
| Topic 10 | 卷, 书, 本, 传, 集, 日, 诗, 文, 余, 度, 志, 撰, 篇, 经, 记 |
| Topic 11 | 命, 官, 贼, 授, 兵, 尔, 巡抚, 营, 阿, 民, 大臣, 部, 明, 馀, 总督 |
| Topic 12 | 州, 官, 诏, 帝, 府, 尚书, 宗, 司, 御史, 王, 郎, 令, 右, 丞, 节度 |
| Topic 13 | 官, 司, 诏, 州, 本, 钱, 令, 路, 据, 民, 奏, 乞, 臣, 军, 差 |
| Topic 14 | 无, 处, 如此, 心, 先生, 须, 如何, 理, 箇, 物, 学, 仁, 他, 底, 却 |
| Topic 15 | 兵, 军, 州, 贼, 遣, 将, 城, 攻, 战, 守, 降, 王, 破, 进, 死 |
| Topic 16 | 县, 州, 水, 置, 东, 山, 南, 西, 郡, 府, 北, 属, 河, 城, 治 |
| Topic 17 | 无, 天下, 民, 能, 不能, 以为, 善, 不可, 行, 生, 者也, 道, 可, 物, 治 |
| Topic 18 | 郑, 大夫, 王, 齐, 诸侯, 音, 传, 伯, 礼, 晋, 释, 无, 服, 书, 侯 |
| Topic 19 | 祭, 礼, 祀, 庙, 官, 乐, 献, 服, 制, 神, 宫, 次, 西, 奏, 设 |

***40 topic model, showing 15 highest probability words in each topic***

| | |
|---|---|
| Topic 0 | 郑, 诸侯, 礼, 音, 释, 王, 正义, 天子, 大夫, 祭, 士, 命, 时, 传, 正 |
| Topic 1 | 歌, 玉, 行, 空, 无, 秋, 归, 飞, 日, 清, 悲, 望, 生, 已, 曲 |
| Topic 2 | 官, 王, 贼, 臣, 都, 廷, 御史, 命, 李, 死, 帝, 忠, 时, 张, 巡抚 |
| Topic 3 | 臣, 陛下, 无, 天下, 已, 议, 可, 伏, 奏, 令, 任, 圣, 国, 朕, 官 |
| Topic 4 | 书, 以为, 时, 无, 不能, 不可, 始, 文, 取, 学, 可, 多, 世, 不知, 唐 |
| Topic 5 | 州, 节度, 王, 刺史, 李, 宗, 军, 诏, 都, 镇, 元, 帝, 唐, 昭, 忠 |
| Topic 6 | 文, 无, 神, 灵, 德, 风, 流, 远, 清, 物, 道, 周, 光, 高, 心 |
| Topic 7 | 為, 於, 無, 國, 軍, 則, 後, 書, 與, 長, 謂, 時, 禮, 諸, 東 |
| Topic 8 | 日, 度, 分, 余, 历, 法, 差, 月, 定, 朔, 减, 行, 加, 半, 数 |
| Topic 9 | 钱, 官, 民, 令, 田, 司, 户, 岁, 盐, 给, 诏, 法, 州, 税, 役 |
| Topic 10 | 卷, 送, 归, 寄, 白, 闲, 应, 山, 题, 客, 居易, 旧, 秋, 春, 日 |
| Topic 11 | 祀, 皇, 帝, 庙, 礼, 祭, 皇帝, 神, 宫, 奏, 宗, 献, 太, 殿, 郊 |
| Topic 12 | 犯, 星, 月, 朔, 岁, 天, 占, 秋, 兵, 日, 夏, 雨, 冬, 太白, 春 |
| Topic 13 | 齐, 晋, 伐, 楚, 伯, 书, 王, 侯, 传, 师, 郑, 叔, 卫, 鲁, 诸侯 |
| Topic 14 | 国, 遣, 王, 遣使, 部, 州, 西, 马, 贡, 地, 夏, 城, 北, 立, 罗 |
| Topic 15 | 卷, 本, 传, 书, 集, 篇, 撰, 志, 录, 文, 经, 按, 引, 记, 宋 |
| Topic 16 | 州, 刺史, 将军, 王, 帝, 封, 令, 尚书, 诏, 孝, 拜, 初, 时, 府, 魏 |
| Topic 17 | 山, 石, 水, 里, 西, 门, 东, 南, 寺, 记, 余, 北, 名, 岩, 木 |
| Topic 18 | 鼓, 金, 服, 乐, 车, 旗, 声, 制, 衣, 律, 黄, 次, 舞, 带, 冠 |
| Topic 19 | 师, 经, 僧, 佛, 时, 道, 无, 寺, 生, 法, 如何, 译, 王, 禅师, 处 |
| Topic 20 | 服, 母, 父, 女, 丧, 葬, 妻, 夫人, 死, 妇, 礼, 亲, 祖, 哭, 无 |
| Topic 21 | 花, 春, 香, 玉, 风, 红, 愁, 似, 更, 月, 烟, 醉, 雨, 酒, 尽 |
| Topic 22 | 诗, 无, 书, 老, 时, 余, 酒, 日, 客, 语, 归, 已, 白, 少, 画 |
| Topic 23 | 命, 尔, 大臣, 授, 馀, 额, 阿, 营, 克, 总督, 部, 明, 巡抚, 布, 乾隆 |
| Topic 24 | 州, 金, 诏, 军, 命, 府, 改作, 王, 都, 路, 司, 赐, 金人, 遣, 宋 |
| Topic 25 | 於, 宾, 西, 拜, 爵, 主人, 引, 升, 位, 官, 前, 设, 再拜, 东, 祝 |
| Topic 26 | 诗, 纪, 赋, 类聚, 御, 览, 引, 集, 切, 汉书, 诗曰, 文选, 初, 乐府, 歌 |
| Topic 27 | 兵, 贼, 军, 将, 攻, 城, 遣, 战, 守, 破, 击, 败, 师, 进, 敌 |
| Topic 28 | 阙, 元, 德, 命, 可, 奉, 无, 赐, 尔, 朝, 伏, 表, 道, 官, 节 |
| Topic 29 | 服, 治, 病, 热, 脉, 汤, 气, 血, 丸, 寒, 升, 钱, 散, 末, 加 |
| Topic 30 | 官, 司, 令, 置, 正, 郎, 掌, 府, 丞, 监, 省, 尚书, 员, 左, 制 |
| Topic 31 | 帝, 迁, 官, 时, 御史, 卒, 尚书, 初, 议, 召, 奏, 诏, 学士, 郎, 吏 |
| Topic 32 | 天下, 无, 民, 能, 不能, 子曰, 善, 治, 行, 者也, 道, 以为, 德, 恶, 而不 |
| Topic 33 | 州, 诏, 本, 司, 官, 臣, 据, 乞, 路, 令, 改, 奏, 阁, 宋, 朝廷 |
| Topic 34 | 处, 心, 无, 如此, 如何, 先生, 箇, 须, 理, 底, 他, 仁, 却, 性, 学 |
| Topic 35 | 王, 侯, 汉, 秦, 帝, 封, 立, 令, 将军, 郡, 魏, 时, 赵, 天下, 吏 |
| Topic 36 | 气, 无, 神, 生, 阳, 象, 阴, 日, 天, 火, 时, 吉, 天地, 形, 阴阳 |
| Topic 37 | 将军, 王, 太守, 刺史, 遣, 郡, 令, 晋, 尚书, 征, 南, 魏, 景, 州, 侯 |
| Topic 38 | 水, 县, 河, 东, 北, 城, 南, 山, 西, 径, 东南, 东北, 郡, 阳, 汉 |
| Topic 39 | 州, 县, 置, 府, 郡, 属, 山, 西, 南, 东, 废, 初, 治, 改, 司 |

***60 topic model, showing 15 highest probability words in each topic***

| | |
|---|---|
| Topic 0 | 州, 诏, 本, 宋, 官, 改, 据, 赐, 殿, 阁, 院, 都, 命, 军, 转运 |
| Topic 1 | 阿, 命, 尔, 哈, 鲁, 路, 赐, 台, 蒙古, 行省, 都, 明, 刺, 中书, 达 |
| Topic 2 | 乐, 舞, 鼓, 服, 金, 声, 律, 制, 宫, 钟, 黄, 车, 奏, 衣, 曲 |
| Topic 3 | 官, 司, 正, 置, 掌, 府, 令, 郎, 监, 丞, 省, 员, 左, 右, 卫 |
| Topic 4 | 官, 御史, 王, 臣, 都, 廷, 忠, 张, 时, 李, 命, 朝, 死, 抚, 府 |
| Topic 5 | 经, 佛, 寺, 僧, 王, 法, 译, 时, 菩萨, 已, 沙门, 名, 身, 无, 善 |
| Topic 6 | 臣, 陛下, 伏, 无, 圣, 任, 已, 奉, 窃, 奏, 望, 谨, 深, 恩, 愿 |
| Topic 7 | 朔, 诏, 夏, 月, 春, 尚书, 秋, 冬, 甲子, 日, 乙卯, 丁卯, 辛卯, 甲申, 庚申 |
| Topic 8 | 軍, 甲, 丙, 國, 部, 會, 政府, 美, 員, 日, 長, 到, 戊, 一二, 委 |
| Topic 9 | 书, 诗, 余, 文, 时, 予, 传, 唐, 元, 学, 多, 名, 世, 古, 少 |
| Topic 10 | 国, 遣, 王, 遣使, 州, 西, 部, 马, 北, 地, 兵, 突厥, 城, 立, 夏 |
| Topic 11 | 州, 县, 郡, 置, 属, 府, 废, 唐, 城, 初, 治, 汉, 改, 领, 晋 |
| Topic 12 | 卷, 集, 撰, 录, 经, 志, 本, 记, 陈氏, 书, 唐, 编, 论, △, 一百 |
| Topic 13 | 帝, 迁, 时, 诏, 官, 奏, 召, 罢, 学士, 议, 进, 卒, 初, 宗, 论 |
| Topic 14 | 风, 无, 高, 流, 玉, 清, 远, 金, 日, 赋, 望, 馀, 光, 游, 山 |
| Topic 15 | 星, 犯, 天, 占, 月, 太白, 兵, 雨, 岁, 荧惑, 赤, 日, 地, 火, 西 |
| Topic 16 | 齐, 晋, 伯, 伐, 书, 楚, 郑, 侯, 传, 师, 卫, 鲁, 叔, 诸侯, 王 |
| Topic 17 | 将军, 刺史, 州, 王, 魏, 景, 齐, 遣, 南, 高祖, 军, 太守, 侯, 都督, 帝 |
| Topic 18 | 命, 大臣, 馀, 总督, 授, 巡抚, 营, 议, 乾隆, 额, 署, 奏, 匪, 康熙, 由 |
| Topic 19 | 為, 無, 於, 則, 國, 時, 與, 後, 書, 見, 長, 來, 將, 漢, 萬 |
| Topic 20 | 為, 禮, 官, 於, 諸, 謂, 議, 奏, 從, 令, 並, 請, 後, 書, 本 |
| Topic 21 | 王, 侯, 秦, 汉, 赵, 臣, 楚, 魏, 将军, 齐, 立, 封, 天下, 国, 兵 |
| Topic 22 | 师, 僧, 如何, 道, 禅师, 处, 无, 和尚, 时, 上堂, 生, 山, 举, 却, 佛 |
| Topic 23 | 服, 汤, 治, 丸, 升, 钱, 末, 加, 右, 味, 散, 取, 煎, 热, 酒 |
| Topic 24 | 迁, 郎, 尚书, 时, 拜, 卒, 令, 年, 郡, 父, 初, 赠, 府, 少, 大夫 |
| Topic 25 | 金, 州, 军, 改作, 府, 金人, 遣, 官, 宗, 宣, 司, 诏, 王, 都, 宋 |
| Topic 26 | 音, 郑, 王, 正义, 诸侯, 释, 天子, 礼, 祭, 掌, 大夫, 命, 地, 官, 朝 |
| Topic 27 | 宾, 於, 主人, 爵, 大夫, 尸, 拜, 礼, 西, 释, 射, 升, 阶, 士, 受 |
| Topic 28 | 州, 节度, 军, 都, 王, 镇, 晋, 刺史, 遣, 契丹, 彦, 帝, 太祖, 全, 行 |
| Topic 29 | 德, 命, 元, 天, 阙, 尔, 皇, 载, 无, 神, 将, 礼, 永, 克, 圣 |
| Topic 30 | 无, 物, 道, 生, 心, 形, 成, 能, 理, 天地, 天, 明, 不能, 性, 可 |
| Topic 31 | 卷, 送, 归, 寄, 应, 白, 春, 树, 山, 秋, 远, 日, 别, 题, 居易 |
| Topic 32 | 他, 就, 和, 介, The, 弟, said, 你, 对, 到, 自己, 旦, Master, 将, 要 |
| Topic 33 | 水, 县, 东, 山, 西, 南, 河, 北, 府, 东南, 东北, 西南, 城, 西北, 志 |
| Topic 34 | 日, 度, 分, 余, 历, 差, 法, 月, 朔, 定, 减, 加, 行, 半, 求 |
| Topic 35 | 阙, 神, 真, 仙, 玄, 灵, 玉, 精, 经, 道, 丹, 元, 宫, 气, 金 |
| Topic 36 | 本, 引, 篇, 按, 传, 校, 汉书, 本作, 御, 误, 文, 音, 书, 史记, 王 |
| Topic 37 | 无, 归, 已, 客, 老, 日, 行, 空, 秋, 酒, 山, 身, 时, 谁, 游 |
| Topic 38 | 州, 节度, 刺史, 李, 御史, 元, 王, 尚书, 宗, 侍郎, 中书, 大夫, 左, 右, 兼 |
| Topic 39 | 死, 时, 令, 无, 因, 家, 将, 数, 杀, 告, 归, 坐, 母, 不能, 已 |
| Topic 40 | 兵, 贼, 军, 将, 攻, 城, 战, 破, 遣, 进, 击, 师, 守, 败, 州 |
| Topic 41 | 生, 白, 食, 木, 鱼, 水, 时, 石, 无, 马, 火, 头, 牛, 草, 多 |
| Topic 42 | 祀, 祭, 庙, 礼, 官, 皇帝, 神, 献, 仪, 郊, 奉, 奏, 引, 殿, 位 |
| Topic 43 | 服, 丧, 父, 葬, 礼, 母, 哭, 祭, 夫人, 祖, 亲, 妇, 大夫, 父母, 士 |
| Topic 44 | 天下, 无, 可, 不可, 以为, 民, 不能, 兵, 固, 国, 能, 制, 将, 已, 易 |
| Topic 45 | 气, 病, 脉, 阴, 阳, 热, 寒, 火, 治, 日, 血, 虚, 伤, 生, 死 |
| Topic 46 | 赋, 切, 诗曰, 音, 汉书, 毛, 善, 子曰, 山, 传, 貌, 左, 志, 余, 诗 |
| Topic 47 | 官, 令, 敕, 朕, 奏, 诏, 已, 制, 举, 司, 可, 任, 刑, 罪, 无 |
| Topic 48 | 旗, 营, 马, 车, 令, 尺, 兵, 总, 军, 前, 寸, 丈, 步, 广, 弓 |
| Topic 49 | 诗, 纪, 类聚, 歌, 本, 乐府, 集, 文选, 文苑, 览, 御, 初, 引, 曲, 玉 |
| Topic 50 | 司, 乞, 官, 诏, 路, 本, 州, 奏, 朝廷, 令, 据, 臣, 元, 已, 安石 |
| Topic 51 | 处, 如此, 箇, 先生, 如何, 心, 须, 仁, 底, 却, 理, 学, 无, 时, 他 |
| Topic 52 | 山, 石, 寺, 记, 里, 西, 南, 门, 东, 岩, 峰, 院, 余, 旧, 文 |
| Topic 53 | 花, 春, 香, 玉, 风, 红, 月, 烟, 愁, 似, 更, 醉, 酒, 梦, 尽 |
| Topic 54 | 钱, 民, 官, 田, 盐, 岁, 户, 给, 令, 税, 石, 州, 司, 诏, 役 |
| Topic 55 | 传, 书, 汉, 周, 帝, 文, 议, 时, 礼, 始, 明, 古, 世, 以为, 王 |
| Topic 56 | 象, 吉, 卦, 无, 乾, 易, 阳, 阴, 贞, 坤, 咎, 正, 位, 刚, 利 |
| Topic 57 | 帝, 王, 封, 皇, 太后, 太子, 孝, 立, 宗, 皇后, 宫, 妃, 皇帝, 薨, 嗣 |
| Topic 58 | 民, 天下, 子曰, 治, 行, 无, 王, 孔子, 善, 能, 臣, 德, 国, 贤, 不能 |
| Topic 59 | 将军, 帝, 太守, 将, 遣, 郡, 王, 侯, 魏, 吴, 晋, 征, 兵, 刺史, 军 |

***80 topic model, showing 15 highest probability words in each topic***

| | |
|---|---|
| Topic 0 | 為, 於, 禮, 官, 諸, 謂, 後, 從, 無, 書, 奏, 令, 則, 議, 請 |
| Topic 1 | 服, 金, 衣, 车, 次, 冠, 制, 旗, 带, 青, 驾, 辂, 饰, 黄, 玉 |
| Topic 2 | 无, 赋, 文, 远, 神, 物, 观, 理, 明, 流, 风, 时, 可, 通, 象 |
| Topic 3 | 州, 诏, 宋, 赐, 殿, 院, 辽, 都, 命, 枢, 学士, 军, 契丹, 宗, 阁 |
| Topic 4 | 州, 县, 置, 府, 属, 郡, 司, 改, 废, 元, 领, 治, 江, 隶, 宁 |
| Topic 5 | 如此, 箇, 如何, 处, 底, 他, 须, 却, 看, 做, 道理, 时, 仁, 都, 先生 |
| Topic 6 | 甲, 丙, 日, 官, 部, 贵, 财, 戊, 旺, 时, 政府, 到, 美, 庚, 运 |
| Topic 7 | 师, 僧, 如何, 道, 禅师, 处, 无, 和尚, 上堂, 山, 生, 时, 举, 却, 你 |
| Topic 8 | 犯, 星, 朔, 月, 岁, 天, 占, 兵, 太白, 日, 甲子, 辛卯, 丁卯, 丙午, 甲寅 |
| Topic 9 | 贼, 兵, 攻, 军, 城, 破, 战, 寇, 败, 州, 进, 将, 陷, 守, 率 |
| Topic 10 | 命, 路, 都, 军, 宋, 赐, 州, 中书, 诏, 行省, 鲁, 司, 剌, 省, 阿 |
| Topic 11 | 州, 帝, 隋, 封, 周, 将军, 总管, 刺史, 王, 太宗, 令, 高祖, 突厥, 府, 都督 |
| Topic 12 | 书, 齐, 传, 侯, 晋, 伯, 郑, 解云, 伐, 师, 称, 卫, 诸侯, 卒, 盟 |
| Topic 13 | 朕, 可, 敕, 诏, 官, 卿, 制, 尔, 赐, 无, 守, 具, 前, 任, 已 |
| Topic 14 | 侯, 帝, 汉, 郡, 将军, 吏, 封, 时, 令, 太守, 传, 王, 诏, 国, 大夫 |
| Topic 15 | 子曰, 孔子, 礼, 孟子, 仁, 行, 义, 孔, 圣人, 无, 章, 天下, 者也, 善, 道 |
| Topic 16 | 迁, 时, 尚书, 卒, 拜, 郎, 初, 令, 少, 卿, 父, 太子, 转, 累, 善 |
| Topic 17 | 无, 臣, 义, 传, 昔, 命, 怀, 实, 志, 德, 国, 可, 令, 兴, 远 |
| Topic 18 | 诗, 唐, 画, 集, 余, 句, 元, 李, 语, 时, 词, 王, 名, 工, 文 |
| Topic 19 | 本, 州, 诏, 据, 司, 官, 路, 令, 改, 馀, 臣, 军, 都, 乞, 卷 |
| Topic 20 | 赋, 汉书, 切, 诗曰, 音, 传, 毛, 子曰, 善, 诗, 记, 山, 汉, 梁, 左 |
| Topic 21 | 齐, 王, 楚, 晋, 臣, 伐, 将, 叔, 国, 吴, 师, 伯, 鲁, 无, 对曰 |
| Topic 22 | 阙, 命, 行, 元, 将, 馀, 道, 朝, 授, 宏, 封, 德, 府, 州, 功 |
| Topic 23 | 臣, 伏, 奉, 表, 谨, 赐, 圣, 无, 恩, 任, 谢, 陛下, 启, 之至, 伏惟 |
| Topic 24 | 州, 节度, 刺史, 李, 元, 宗, 尚书, 侍郎, 御史, 王, 中书, 诏, 右, 朝, 左 |
| Topic 25 | 行, 思, 无, 怀, 心, 悲, 生, 哀, 远, 余, 志, 游, 世, 将, 尔 |
| Topic 26 | 山, 游, 风, 清, 日, 已, 望, 林, 归, 树, 月, 空, 初, 秋, 开 |
| Topic 27 | 西, 官, 拜, 位, 引, 爵, 前, 升, 再拜, 设, 东, 仪, 阶, 主人, 立 |
| Topic 28 | 服, 丧, 礼, 父, 哭, 大夫, 士, 葬, 冠, 母, 祭, 小, 无, 祖, 衰 |
| Topic 29 | 民, 天下, 治, 国, 刑, 无, 政, 法, 行, 善, 王, 明, 乱, 令, 臣 |
| Topic 30 | 本, 按, 引, 篇, 本作, 桉, 传, 文, 误, 据, 校, 补, 汉书, 王, 御 |
| Topic 31 | 為, 軍, 國, 無, 於, 長, 則, 與, 時, 東, 見, 後, 來, 將, 會 |
| Topic 32 | 臣, 陛下, 天下, 议, 奏, 无, 朝廷, 谏, 以为, 恐, 不可, 圣, 已, 窃, 伏 |
| Topic 33 | 卷, 送, 寄, 白, 归, 居易, 闲, 应, 远, 题, 秋, 客, 无, 春, 尽 |
| Topic 34 | 祭, 祀, 庙, 礼, 郊, 神, 配, 祠, 坛, 祖, 享, 帝, 议, 献, 宗 |
| Topic 35 | 帝, 官, 御史, 卒, 诏, 迁, 民, 时, 吏, 治, 召, 议, 奏, 罢, 擢 |
| Topic 36 | 水, 东, 县, 山, 西, 南, 北, 河, 东南, 东北, 西南, 西北, 径, 府, 志 |
| Topic 37 | 佛, 寺, 经, 僧, 时, 王, 法, 无, 已, 身, 善, 菩萨, 名, 生, 受 |
| Topic 38 | 尔, 阿, 明, 布, 额, 哈, 命, 图, 喇, 克, 特, 勒, 授, 馀, 喀 |
| Topic 39 | 官, 臣, 王, 都, 御史, 廷, 忠, 张, 时, 李, 命, 巡抚, 已, 左, 朝 |
| Topic 40 | 馀, 命, 巡抚, 总督, 大臣, 议, 署, 匪, 由, 调, 乾隆, 奏, 免, 州县, 设 |
| Topic 41 | 书, 传, 史, 文, 志, 汉, 论, 古, 春秋, 篇, 记, 经, 序, 着, 撰 |
| Topic 42 | 无, 归, 行, 客, 生, 歌, 白, 空, 送, 时, 已, 身, 天, 老, 谁 |
| Topic 43 | 德, 皇, 神, 圣, 天, 灵, 献, 降, 乐, 肃, 和, 帝, 礼, 昭, 载 |
| Topic 44 | 兵, 将, 遣, 军, 城, 攻, 骑, 战, 击, 馀, 无, 数, 降, 斩, 因 |
| Topic 45 | 钱, 民, 田, 官, 盐, 岁, 户, 税, 石, 给, 州, 令, 诏, 司, 役 |
| Topic 46 | 象, 吉, 无, 卦, 阳, 乾, 易, 阴, 坤, 贞, 正, 咎, 位, 物, 凶 |
| Topic 47 | 气, 病, 脉, 热, 治, 寒, 阴, 血, 阳, 虚, 痛, 火, 邪, 补, 证 |
| Topic 48 | 将军, 晋, 魏, 帝, 王, 吴, 太守, 将, 蜀, 刺史, 郡, 亮, 司马, 败, 征 |
| Topic 49 | 乞, 元, 司, 安石, 熙, 朝廷, 罢, 除, 祐, 宗, 官, 进, 圣, 诏, 御史 |
| Topic 50 | 金, 州, 改作, 金人, 军, 府, 诏, 宣, 遣, 兵, 司, 淮, 兼, 宗, 王 |
| Topic 51 | 帝, 王, 皇, 太后, 封, 太子, 宗, 皇后, 立, 孝, 宫, 皇帝, 殿, 诏, 妃 |
| Topic 52 | 无, 物, 道, 生, 形, 成, 能, 天下, 不能, 不知, 德, 天地, 音, 而不, 名 |
| Topic 53 | 死, 时, 母, 女, 家, 妻, 无, 父, 因, 妇, 生, 归, 杀, 食, 语 |
| Topic 54 | 卷, 经, 集, 录, 撰, 译, 本, 纸, 右, △, 陈氏, 一百, 第, 编, 论 |
| Topic 55 | 释, 郑, 大夫, 於, 礼, 宾, 射, 诸侯, 士, 王, 音, 天子, 祭, 命, 司 |
| Topic 56 | 心, 无, 性, 理, 物, 气, 先生, 学, 所谓, 善, 体, 天地, 发, 动, 静 |
| Topic 57 | 正义, 王, 周, 音, 传, 命, 殷, 伯, 德, 姓, 文, 诗, 夏, 舜, 生 |
| Topic 58 | 天下, 无, 可, 不能, 固, 不可, 以为, 民, 兵, 能, 者也, 国, 已, 至于, 而不 |
| Topic 59 | 乐, 舞, 声, 鼓, 律, 歌, 奏, 宫, 钟, 曲, 锺, 应, 音, 弦, 调 |
| Topic 60 | 官, 司, 正, 置, 掌, 府, 令, 丞, 郎, 监, 员, 省, 左, 右, 史 |
| Topic 61 | 石, 山, 里, 门, 余, 西, 寺, 院, 记, 南, 洞, 东, 岩, 峰, 桥 |
| Topic 62 | 官, 举, 学士, 进士, 制, 郎, 考, 司, 诏, 选, 科, 学, 士, 补, 文 |
| Topic 63 | 国, 王, 遣使, 西, 遣, 部, 贡, 地, 北, 马, 城, 夏, 立, 南, 居 |
| Topic 64 | 令, 官, 奏, 已, 司, 敕, 准, 罪, 应, 本, 须, 不得, 差, 合, 法 |
| Topic 65 | 服, 升, 钱, 汤, 丸, 治, 末, 味, 加, 右, 取, 酒, 甘草, 煎, 每 |
| Topic 66 | 书, 余, 已, 可, 予, 无, 少, 时, 不能, 多, 日, 意, 行, 尔, 能 |
| Topic 67 | 将军, 王, 刺史, 太守, 州, 景, 侯, 南, 郡, 军, 遣, 义, 梁, 参军, 领 |
| Topic 68 | 花, 春, 醉, 香, 酒, 雨, 梦, 更, 似, 风, 尽, 闲, 老, 红, 小 |
| Topic 69 | 县, 郡, 州, 置, 城, 属, 汉, 晋, 初, 水, 治, 阳, 废, 唐, 河 |
| Topic 70 | 王, 秦, 侯, 赵, 楚, 魏, 齐, 汉, 臣, 天下, 燕, 兵, 立, 诸侯, 击 |
| Topic 71 | 州, 节度, 军, 都, 王, 镇, 刺史, 晋, 彦, 契丹, 全, 太祖, 唐, 指挥, 延 |
| Topic 72 | 神, 真, 仙, 玄, 精, 经, 气, 道, 灵, 玉, 丹, 金, 天, 生, 宫 |
| Topic 73 | 将军, 刺史, 州, 魏, 王, 尚书, 诏, 爵, 除, 都督, 武, 元, 侯, 仪, 齐 |
| Topic 74 | 日, 度, 分, 余, 历, 月, 差, 法, 朔, 减, 定, 行, 加, 半, 求 |
| Topic 75 | 诗, 纪, 类聚, 览, 御, 引, 文选, 文苑, 乐府, 集, 本, 学记, 初, 艺文, 十八 |
| Topic 76 | 玉, 花, 歌, 曲, 金, 春, 香, 飞, 风, 罗, 月, 夜, 轻, 树, 愁 |
| Topic 77 | 铭, 夫人, 讳, 年, 州, 缺, 碑, 志, 父, 德, 文, 生, 葬, 府, 世 |
| Topic 78 | 生, 木, 雨, 水, 火, 时, 食, 白, 气, 土, 民, 旱, 日, 赤, 阴 |
| Topic 79 | 兵, 军, 营, 敌, 旗, 马, 将, 阵, 总, 战, 令, 车, 备, 贼, 队 |

***100 topic model, showing 15 highest probability words in each topic***

| | |
|---|---|
| Topic 0 | 本, 按, 引, 篇, 传, 校, 本作, 汉书, 误, 史记, 文, 御, 据, 校, 书 |
| Topic 1 | 春, 花, 香, 醉, 愁, 红, 梦, 风, 清, 更, 烟, 玉, 谁, 尽, 前 |
| Topic 2 | 晋, 帝, 传, 魏, 汉, 志, 令, 武帝, 王, 诏, 始, 书, 吴, 元, 文帝 |
| Topic 3 | 州, 府, 蛮, 司, 江, 卫, 置, 土, 属, 地, 元, 寨, 改, 川, 官司 |
| Topic 4 | 天下, 子曰, 民, 孔子, 孟子, 治, 者也, 善, 仁, 孔, 不能, 王, 而不, 章, 义 |
| Topic 5 | 為, 謂, 本, 罪, 減, 官, 議, 從, 徒, 加, 錢, 殺, 無, 諸, 應 |
| Topic 6 | 乐, 舞, 声, 鼓, 律, 奏, 歌, 宫, 钟, 曲, 锺, 弦, 音, 调, 羽 |
| Topic 7 | 国, 王, 遣使, 遣, 西, 部, 贡, 北, 地, 南, 立, 城, 东, 可汗, 居 |
| Topic 8 | 州, 县, 郡, 置, 属, 领, 废, 治, 汉, 户, 改, 府, 城, 初, 阳 |
| Topic 9 | 音, 切, 木, 鱼, 食, 生, 无, 实, 水, 白, ?, 多, 名, 似, 谓之 |
| Topic 10 | 时, 身, 生, 无, 已, 令, 心, 须, 食, 常, 多, 不得, 语, 能, 念 |
| Topic 11 | 兵, 军, 将, 敌, 战, 胜, 贼, 备, 可, 计, 卒, 利, 攻, 守, 虏 |
| Topic 12 | 赋, 诗曰, 汉书, 毛, 子曰, 善, 传, 切, 诗, 左, 貌, 尚书, 楚辞, 说文, 礼记 |
| Topic 13 | 歌, 行, 无, 归, 空, 胡, 送, 悲, 天, 辞, 白, 生, 谁, 赠, 愁 |
| Topic 14 | 师, 僧, 如何, 道, 禅师, 处, 无, 和尚, 山, 上堂, 生, 时, 举, 却, 佛 |
| Topic 15 | 将军, 刺史, 王, 太守, 州, 南, 景, 军, 遣, 义, 郡, 高祖, 领, 参军, 侯 |
| Topic 16 | 宋, 命, 军, 路, 都, 赐, 行省, 刺, 中书, 鲁, 阿, 政事, 里, 府, 金 |
| Topic 17 | 迁, 卒, 官, 进士, 诏, 时, 授, 初, 改, 举, 擢, 尚书, 召, 年, 进 |
| Topic 18 | 晋, 伯, 楚, 齐, 郑, 叔, 伐, 师, 侯, 王, 鲁, 卫, 季, 正义, 国 |
| Topic 19 | 神, 皇, 德, 圣, 乐, 献, 歌, 天, 灵, 奏, 宫, 礼, 和, 肃, 昭 |
| Topic 20 | 服, 升, 钱, 汤, 丸, 末, 治, 味, 加, 右, 酒, 取, 生, 甘草, 每 |
| Topic 21 | 民, 田, 令, 吏, 无, 治, 官, 多, 国, 岁, 役, 赋, 农, 地, 天下 |
| Topic 22 | 天下, 无, 固, 可, 不能, 民, 不可, 国, 已, 者也, 亡, 恶, 乱, 权, 天子 |
| Topic 23 | 花, 玉, 春, 夜, 曲, 秋, 风, 金, 月, 开, 树, 水, 对, 叶, 飞 |
| Topic 24 | 宾, 於, 主人, 爵, 尸, 大夫, 西, 拜, 礼, 释, 射, 阶, 升, 祭, 东 |
| Topic 25 | 朕, 可, 敕, 赐, 卿, 尔, 官, 制, 守, 具, 节, 授, 元, 诏, 俾 |
| Topic 26 | 刑, 罪, 狱, 法, 盗, 赦, 令, 吏, 律, 杖, 官, 杀, 犯, 断, 死 |
| Topic 27 | 如此, 箇, 如何, 底, 处, 须, 他, 却, 看, 做, 仁, 道理, 都, 时, 理会 |
| Topic 28 | 节度, 州, 军, 兵, 贼, 诏, 将, 宗, 光, 忠, 李, 全, 王, 怀, 武 |
| Topic 29 | 死, 母, 家, 时, 父, 女, 妻, 因, 归, 年, 卒, 无, 门, 数, 日 |
| Topic 30 | 尔, 明, 布, 阿, 哈, 额, 图, 克, 喇, 兵, 特, 勒, 命, 授, 师 |
| Topic 31 | 无, 行, 将, 道, 礼, 可, 辞, 能, 不能, 善, 德, 文, 信, 以为, 不可 |
| Topic 32 | 无, 德, 命, 实, 庶, 天, 已, 王, 皇, 礼, 明, 怀, 典, 戎, 图 |
| Topic 33 | 臣, 帝, 陛下, 奏, 议, 罢, 论, 谏, 无, 朝廷, 天下, 时, 召, 大臣, 因 |
| Topic 34 | 卷, 集, 撰, 录, 陈氏, 志, 本, 记, 一百, △, 编, 唐, 经, 图, 右 |
| Topic 35 | 臣, 乞, 司, 官, 诏, 奏, 朝廷, 安石, 路, 元, 已, 令, 差, 行, 除 |
| Topic 36 | 州, 刺史, 御史, 侍郎, 元, 节度, 李, 中书, 尚书, 大夫, 宗, 平章, 制, 崔, 右 |
| Topic 37 | 石, 山, 余, 西, 寺, 门, 岩, 峰, 洞, 桥, 亭, 南, 里, 舟, 水 |
| Topic 38 | 王, 帝, 封, 皇, 太后, 宗, 太子, 立, 孝, 皇后, 妃, 薨, 嗣, 宫, 谥 |
| Topic 39 | 雨, 州, 旱, 地震, 民, 饥, 大水, 雹, 火, 木, 坏, 秋, 生, 蝗, 七年 |
| Topic 40 | 臣, 伏, 奉, 陛下, 谨, 圣, 无, 表, 恩, 任, 谢, 赐, 启, 蒙, 状 |
| Topic 41 | 书, 传, 文, 学, 史, 经, 春秋, 篇, 义, 着, 儒, 易, 序, 汉, 论 |
| Topic 42 | 水, 东, 县, 北, 径, 南, 山, 西, 城, 阳, 汉, 河, 东北, 谓之, 东南 |
| Topic 43 | 将军, 刺史, 州, 魏, 王, 诏, 除, 爵, 尚书, 元, 都督, 散, 赠, 骑, 节 |
| Topic 44 | 真, 神, 仙, 玄, 经, 丹, 灵, 玉, 精, 道, 气, 元, 宫, 金, 真人 |
| Topic 45 | 诏, 州, 殿, 学士, 院, 赐, 命, 制, 朝, 官, 枢, 辽, 宗, 阁, 臣 |
| Topic 46 | 服, 丧, 父, 礼, 哭, 葬, 母, 大夫, 冠, 士, 祖, 父母, 小, 衰, 祭 |
| Topic 47 | 侯, 汉, 将军, 王, 匈奴, 时, 吏, 丞相, 封, 大夫, 郡, 令, 孝, 单于, 莽 |
| Topic 48 | 郑, 释, 诸侯, 音, 王, 天子, 大夫, 礼, 掌, 朝, 士, 桉, 司, 正义, 卿 |
| Topic 49 | 县, 州, 置, 城, 初, 志, 郡, 属, 府, 宋, 西, 唐, 废, 山, 治 |
| Topic 50 | 铭, 州, 夫人, 讳, 府, 缺, 生, 年, 德, 志, 碑, 文, 先, 父, 郡 |
| Topic 51 | 病, 气, 脉, 热, 治, 寒, 阴, 血, 阳, 虚, 痛, 火, 邪, 补, 证 |
| Topic 52 | 帝, 将军, 郡, 侯, 太守, 汉, 拜, 封, 将, 吏, 令, 书, 时, 初, 征 |
| Topic 53 | 卷, 送, 归, 寄, 白, 山, 居易, 远, 题, 别, 应, 闲, 吟, 客, 无 |
| Topic 54 | 王, 秦, 楚, 齐, 赵, 魏, 臣, 燕, 韩, 天下, 侯, 吴, 兵, 攻, 立 |
| Topic 55 | 官, 臣, 御史, 帝, 王, 都, 时, 廷, 命, 忠, 朝, 李, 张, 文, 尚书 |
| Topic 56 | 尚书, 迁, 时, 郎, 令, 初, 议, 拜, 卿, 转, 官, 少, 除, 太子, 大夫 |
| Topic 57 | 河, 水, 决, 湖, 淮, 堤, 开, 口, 筑, 闸, 浚, 堰, 运, 东, 通 |
| Topic 58 | 心, 无, 先生, 性, 学, 善, 理, 物, 所谓, 恶, 发, 体, 气, 致, 动 |
| Topic 59 | 本, 据, 诏, 宋, 改, 州, 卷, 馀, 官, 司, 阁, 觿, 令, 书, 原 |
| Topic 60 | 石, 马, 白, 龙, 山, 地, 生, 时, 金, 天, 木, 鸟, 火, 鱼, 象 |
| Topic 61 | 祭, 祀, 庙, 礼, 郊, 神, 配, 祖, 祠, 享, 坛, 宗, 献, 议, 帝 |
| Topic 62 | 贼, 营, 兵, 总兵, 镇, 军, 寇, 剿, 馀, 克, 陷, 砲, 匪, 擢, 国 |
| Topic 63 | 命, 大臣, 巡抚, 总督, 奏, 议, 学士, 谕, 乾隆, 署, 由, 授, 州县, 直, 调 |
| Topic 64 | 州, 节度, 军, 都, 帝, 晋, 契丹, 彦, 太祖, 镇, 王, 刺史, 唐, 延, 指挥 |
| Topic 65 | 改作, 金人, 金, 州, 军, 府, 虏, 遣, 兵, 宣, 统制, 敌, 行, 司, 王 |
| Topic 66 | 道, 理, 世, 俗, 体, 可, 实, 德, 论, 莫, 形, 情, 难, 远, 殊 |
| Topic 67 | 王, 命, 民, 周, 德, 传, 舜, 殷, 尧, 天, 禹, 帝, 正义, 周公, 文王 |
| Topic 68 | 日, 度, 分, 余, 历, 差, 法, 朔, 减, 月, 定, 加, 行, 半, 求 |
| Topic 69 | 东, 山, 南, 西, 北, 水, 县, 府, 溪, 东南, 江, 西南, 河, 东北, 合 |
| Topic 70 | 服, 金, 衣, 车, 冠, 旗, 带, 制, 次, 青, 辂, 饰, 驾, 黄, 玉 |
| Topic 71 | 為, 於, 無, 則, 國, 書, 後, 禮, 與, 時, 東, 長, 將, 軍, 見 |
| Topic 72 | 州, 帝, 刺史, 齐, 武, 侯, 仪, 封, 梁, 景, 周, 魏, 孝, 开, 府 |
| Topic 73 | 诗, 纪, 类聚, 览, 御, 引, 文选, 文苑, 乐府, 本, 集, 学记, 初, 十九, 十八 |
| Topic 74 | 旗, 总, 营, 队, 城, 令, 尺, 车, 丈, ⊙, 每, 俱, 枪, The, 步 |
| Topic 75 | 官, 引, 殿, 皇帝, 仪, 位, 前, 再拜, 西, 次, 奉, 设, 诣, 立, 跪 |
| Topic 76 | 诏, 朔, 尚书, 王, 夏, 州, 春, 秋, 皇, 冬, 幸, 遣使, 罢, 甲子, 宫 |
| Topic 77 | 游, 清, 无, 望, 日, 时, 怀, 山, 思, 风, 远, 高, 悲, 流, 心 |
| Topic 78 | 无, 老, 酒, 已, 客, 日, 归, 秋, 时, 雨, 山, 初, 闲, 夜, 书 |
| Topic 79 | 以为, 无, 不能, 不可, 可, 天下, 不知, 至于, 所谓, 古, 取, 尽, 已, 固, 书 |
| Topic 80 | 文, 金, 置, 绍兴, 钱, 宗, 院, 熙, 崇, 诏, 兼, 旧, 元, 宁, 州 |
| Topic 81 | 星, 犯, 月, 日, 占, 太白, 天, 岁, 荧惑, 火, 辰, 官, 兵, 气, 贵 |
| Topic 82 | 无, 物, 音, 成, 形, 不知, 道, 篇, 天下, 宣, 司, 名, 游, 邪, 德 |
| Topic 83 | 钱, 盐, 官, 户, 石, 给, 岁, 司, 税, 米, 州, 诏, 贯, 纳, 铸 |
| Topic 84 | 书, 画, 图, 笔, 纸, 帖, 王, 弟, 工, 草, 余, 善, 墨, 法, 写 |
| Topic 85 | 官, 奏, 令, 敕, 已, 准, 司, 举, 本, 任, 应, 州, 须, 不得, 委 |
| Topic 86 | 生, 天, 无, 物, 天地, 道, 气, 明, 地, 形, 万物, 圣人, 神, 能, 成 |
| Topic 87 | 书, 齐, 解云, 传, 侯, 称, 伐, 是也, 伯, 卒, 大夫, 诸侯, 夏, 文, 葬 |
| Topic 88 | 兵, 贼, 遣, 城, 攻, 将, 军, 降, 破, 战, 进, 守, 击, 斩, 率 |
| Topic 89 | 阙, 元, €, 馀, 将, 因, 赋, 二字, 观, 宏, 无, 金, 玉, 时, 居 |
| Topic 90 | 将军, 王, 遣, 将, 刺史, 帝, 太守, 魏, 镇, 秦, 晋, 慕容, 讨, 坚, 玄 |
| Topic 91 | 臣, 天下, 陛下, 无, 国, 明, 以为, 失, 乱, 贤, 不可, 行, 安, 不能, 谏 |
| Topic 92 | 州, 军, 边, 兵, 夏, 路, 都, 河, 寨, 蕃, 西, 庆, 马, 保, 指挥 |
| Topic 93 | 州, 帝, 隋, 太宗, 总管, 令, 将军, 突厥, 王, 封, 刺史, 高祖, 左, 太子, 玄 |
| Topic 94 | 象, 卦, 吉, 无, 乾, 贞, 坤, 咎, 阳, 易, 正, 位, 阴, 刚, 利 |
| Topic 95 | 官, 置, 掌, 令, 丞, 府, 司, 郎, 正, 监, 右, 左, 将军, 卫, 曹 |
| Topic 96 | 经, 佛, 寺, 僧, 法, 译, 菩萨, 本, 沙门, 王, 释, 名, 塔, 录, 部 |
| Topic 97 | 诗, 余, 予, 文, 时, 语, 句, 元, 李, 书, 唐, 集, 词, 名, 诗云 |
| Topic 98 | 軍, 甲, 丙, 國, 部, 會, 政府, 美, 日, 到, 戊, 共, 一二, 員, 上海 |
| Topic 99 | 司, 官, 正, 员, 省, 置, 都, 监, 内, 掌, 副, 郎, 院, 领, 职 |

**Appendix 6:** Distance measures

One of the strengths of LDA topic modeling lies in the fact that documents and topics are represented within the same probability space, making it possible to apply standard information theoretic measures to compute document-document, document-topic, and topic-topic similarities.

Jensen-Shannon Distance (JSD) (Endres & Schindelin 2003) is a widely used metric from information theory which measures the difference in bits between two probability distributions.

For this project, we computed the similarity between two documents as 1-JSD(P,Q), where JSD(P,Q) is the Jensen-Shannon Distance between the probability distributions corresponding to the topic mixtures P and Q of the documents.

Jensen-Shannon Distance is defined as the square root of the Jensen-Shannon *divergence* between the distributions P and Q.

Jensen-Shannon divergence in turn is based upon averaging the Kullback-Leibler (KL) divergences between the distributions (Kullback & Leibler 1951).

Informally, KL measures the degree to which observations drawn from Q are unexpected when the expected values are given by P. KL is not symmetric: what is surprising about Q from the perspective of P needs not be the same as what is surprising about P from the perspective of Q.

Precise definitions of these measures can be found in the works cited and at https://en.wikipedia.org/wiki/Kullback–Leibler_divergence and https://en.wikipedia.org/wiki/Jensen–Shannon_divergence.

For topic-mediated queries we first compute word-topic similarity by creating a “pseudo-document” which distributes the available probability across the query terms. (In the case of the example in the main text of a query using the single term for yinyang, this means that the probability assigned to that term is 1, and all other terms are assigned 0.) This pseudo topic is then used to compute similarities to the model’s topics using the JSD-based similarity function described above. A new topic distribution vector, weighted according to the pseudo-document’s similarity to each of the topic-word distributions, is used to compute the nearest actual documents using the JSD-based similarity measure.